\documentclass{article}

\usepackage[main,final]{neurips_2026}

\usepackage[utf8]{inputenc} % allow utf-8 input
\usepackage[T1]{fontenc}    % use 8-bit T1 fonts
\usepackage{hyperref}       % hyperlinks
\hypersetup{hidelinks}
\usepackage{url}            % simple URL typesetting
\usepackage{booktabs}       % professional-quality tables
\usepackage{amsfonts}       % blackboard math symbols
\usepackage{nicefrac}       % compact symbols for 1/2, etc.
\usepackage{microtype}      % microtypography
\usepackage{xcolor}         % colors
\usepackage{amsmath} 
\usepackage{amsthm}
\usepackage{algorithm} 
\usepackage{algorithmic}

\usepackage{multirow}
\usepackage{graphicx}
\usepackage{booktabs}    
\usepackage{subcaption}  
\usepackage{caption}       
\usepackage{wrapfig}
\usepackage{caption}

\title{Closed-Form Linear-Probe Dataset Distillation for Pre-trained Vision Models}

\author{%
  Bincheng Peng $^{\dagger}$ \\
  Hokkaido University \\
  \texttt{peng@lmd.ist.hokudai.ac.jp}
  \And
  Guang Li $^{\dagger,*}$ \\
  Hokkaido University \\
  \texttt{guang@lmd.ist.hokudai.ac.jp}
  \AND
  Ping Liu \\
  University of Nevada, Reno \\
  \texttt{pingl@unr.edu}
  \And
  Takahiro Ogawa \\
  Hokkaido University \\
  \texttt{ogawa@lmd.ist.hokudai.ac.jp}
  \AND
  Miki Haseyama \\
  Hokkaido University \\
  \texttt{mhaseyama@lmd.ist.hokudai.ac.jp}
}

\begin{document}

\maketitle

\begingroup
\renewcommand{\thefootnote}{}
\footnotetext{$^{\dagger}$ Equal contribution.}
\footnotetext{$^{*}$ Correspondence to: Guang Li
<guang@lmd.ist.hokudai.ac.jp>.}
\endgroup

\begin{abstract}
Dataset distillation compresses a large training set into a small synthetic set that preserves downstream training utility.
While most existing methods target training networks from scratch, modern visual transfer learning often uses frozen pre-trained encoders followed by lightweight linear probing.
Existing distillation methods for this setting either rely on linear-probe gradient matching or use kernel and feature-regression formulations originally designed for from-scratch training.
Frozen-feature linear probing admits a closed-form solution determined directly by the pre-trained features themselves, with no infinite-width approximation and no inner-loop trajectory.
We propose Closed-Form Linear-Probe Dataset Distillation (CLP-DD), a bilevel formulation that computes the linear probe induced by the synthetic set with a sample-space kernel ridge solver. The synthetic images are then updated by evaluating this induced classifier on real features through a temperature-scaled softmax cross-entropy, where the classifier columns act as learned class anchors in feature space.
We further show that the choice of outer objective is decisive: replacing the class-anchor objective with a standard MSE outer loss reduces average accuracy by 2.9 percentage points on ImageNet-100.
On ImageNet-100 with four pre-trained backbones, CLP-DD substantially improves over LGM without DSA and approaches LGM with DSA at a fraction of the computational cost.
On ImageNet-1K, CLP-DD achieves accuracy comparable to LGM with DSA; profiling with DINOv2 at IPC$=$1 shows roughly $14\times$ faster optimization and less than one-eighth of the peak GPU memory.
The code is available at \url{https://github.com/Guang000/CLP-DD}.
\end{abstract}

\section{Introduction}
Large-scale pre-training has reshaped visual transfer learning. Modern pipelines commonly use powerful pre-trained vision models, such as CLIP~\cite{radford2021learning}, DINOv2~\cite{oquab2024dinov2}, MoCo-v3~\cite{chen2021empirical}, and EVA-02~\cite{fang2023eva}, as frozen feature extractors followed by a lightweight linear classifier. In this regime, dataset distillation (DD)~\cite{wang2018dataset, li2022awesome, yu2023review} seeks to compress a large training set into a small synthetic set that preserves downstream linear-probe performance. As pre-trained backbones continue to grow, full-network distillation becomes increasingly expensive, making frozen-backbone linear probing a natural setting in which DD can scale. This setting inherits the strong representations of large pre-trained encoders while making the inner adaptation problem simple enough to admit structural acceleration.

Existing DD methods relevant to this setting fall into two broad families, neither of which fully exploits the structure available once the encoder is fixed. Gradient- and trajectory-matching methods match classifier gradients or training trajectories; the recent Linear Gradient Matching (LGM)~\cite{cazenavette2025dataset} matches linear-probe gradients in the frozen-backbone setting, and obtains its strongest results when combined with Differentiable Siamese Augmentation (DSA)~\cite{dsa}, at substantial computational cost. Kernel-based closed-form methods such as KIP~\cite{nguyen2020dataset} and FRePo~\cite{zhou2022dataset} replace the inner trajectory with a kernel ridge regression solution. Both were developed primarily for from-scratch training: KIP uses NTK-based regression, whereas FRePo uses neural feature regression with a pool of evolving models. Once the encoder is fixed, however, the inner problem reduces to a ridge-regularized linear solve on pre-trained features: the relevant kernel is simply the empirical Gram matrix of frozen features, and no NTK approximation is required. This structural simplification suggests a more direct route to frozen-backbone DD than costly gradient matching or DSA-heavy optimization.

\begin{wrapfigure}{r}{0.45\textwidth}
    \vspace{-4mm}
    \centering
    \includegraphics[width=\linewidth]{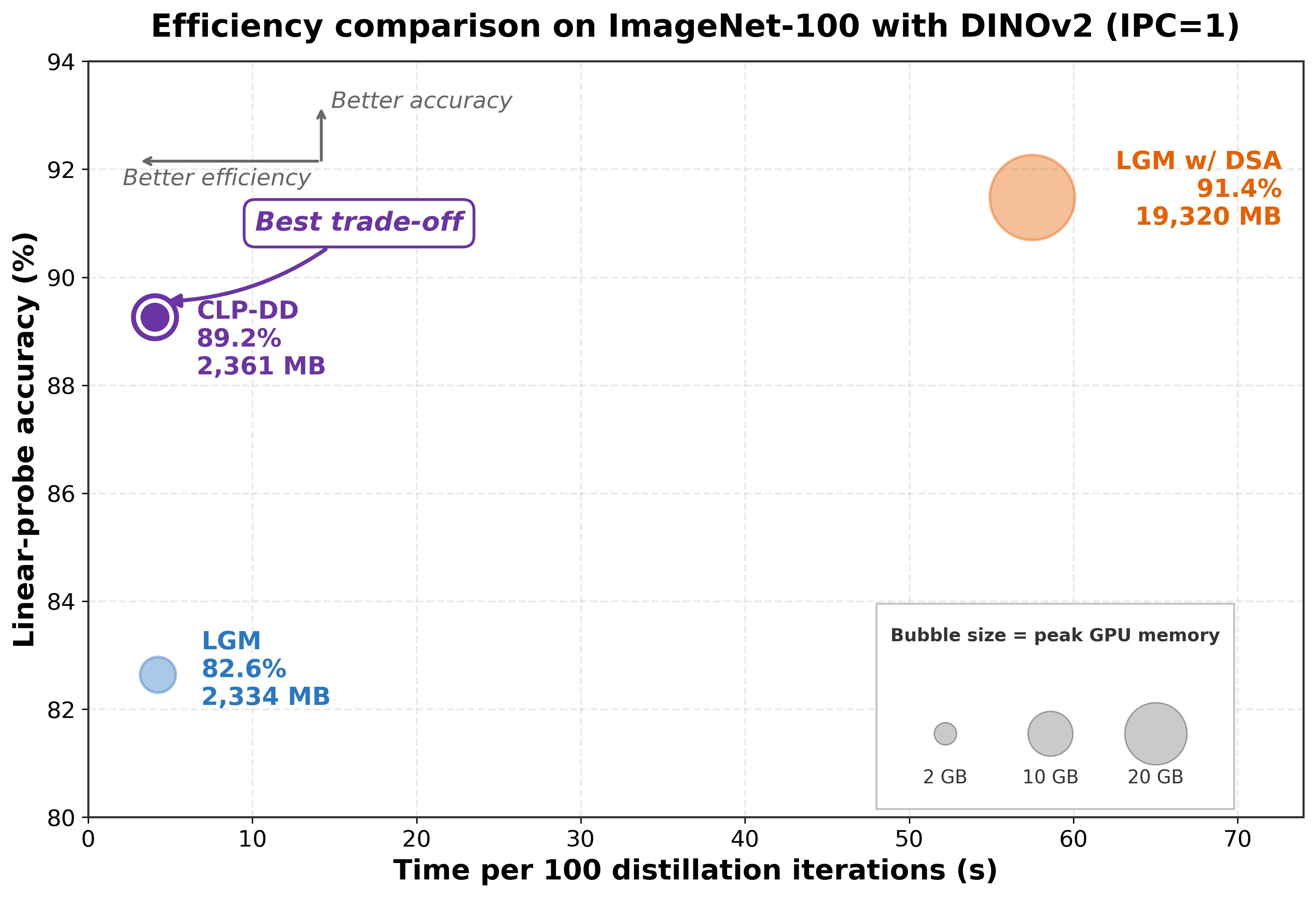}
    \caption{Accuracy versus efficiency on ImageNet-100 with DINOv2 at IPC$=$1. CLP-DD attains accuracy close to LGM with DSA at the runtime and memory of LGM without DSA. Bubble area denotes peak GPU memory.}
    \label{fig:intro_efficiency}
    \vspace{-4mm}
\end{wrapfigure}

We propose Closed-Form Linear-Probe Dataset Distillation (CLP-DD), a bilevel framework that turns this structural simplification into a concrete distillation algorithm. The inner adaptation problem in our setting is structurally simpler than in from-scratch DD: it is a strongly convex ridge regression on fixed features rather than a non-convex deep-network optimization, and it has a single closed-form minimizer that does not depend on initialization, learning rate, or the number of inner steps. In the low-IPC regime where the number of synthetic samples $N$ is much smaller than the feature dimension $d$, this minimizer can equivalently be written as a sample-space kernel ridge regression, replacing a $d \times d$ inverse with an $N \times N$ inverse while preserving differentiability through the solve end-to-end.

Building on this closed-form inner step, we further design the outer objective to align synthetic-image optimization with linear-probe transfer. We find that pairing this solver with the standard ridge MSE loss, which corresponds to a frozen-backbone analogue of KIP, underperforms the class-anchor outer objective. The bottleneck is that MSE penalizes squared deviation from one-hot targets uniformly across class dimensions, whereas the linear-probe transfer task only requires correct class discrimination on real features. CLP-DD therefore evaluates the induced classifier on a small class-balanced batch of real features through a temperature-scaled softmax cross-entropy, where the columns of the closed-form classifier act as learned class anchors in feature space. This aligns the synthetic-image gradient with the actual linear-probe evaluation objective rather than an MSE surrogate, while requiring neither trajectory unrolling nor heavy differentiable augmentation like DSA.

To validate this design, we evaluate CLP-DD on ImageNet-100 and ImageNet-1K with four pre-trained backbones at IPC$=$1, demonstrating strong performance without DSA and accuracy comparable to LGM with DSA on ImageNet-1K. As shown in Figure~\ref{fig:intro_efficiency}, CLP-DD attains this accuracy with the runtime and memory of DSA-free LGM, about $14\times$ faster and using less than one-eighth of the peak GPU memory of DSA-based LGM, providing a favorable accuracy--efficiency balance on the frontier of low-IPC frozen-backbone dataset distillation.

Our contributions are summarized as follows:
\begin{itemize}
    \item We identify a structural opportunity in frozen-backbone dataset distillation: frozen-feature linear probing admits an exact closed-form solution determined directly by the pre-trained features. In the low-IPC regime, this solution can be computed in sample-space kernel form, avoiding both inner-loop unrolling and NTK-based kernel approximation.

    \item We identify a discriminative-outer-objective gap for closed-form distillation with frozen backbones. Pairing the closed-form inner solver with a temperature-scaled softmax cross-entropy objective treats its classifier columns as class anchors and improves average accuracy over an MSE outer loss, a frozen-backbone analogue of KIP.

    \item We propose CLP-DD, which combines the sample-space closed-form solver with the discriminative outer objective. On ImageNet-100 and ImageNet-1K with four pre-trained backbones, CLP-DD improves average accuracy over real-image baselines and DSA-free LGM, and achieves accuracy comparable to LGM with DSA on ImageNet-1K. DINOv2 profiling at IPC$=$1 shows roughly $14\times$ faster optimization and less than one-eighth of the peak GPU memory of DSA-based LGM.
\end{itemize}

\section{Related Work}
\label{sec:related}

\textbf{Dataset Distillation.}
Dataset distillation (DD) aims to compress a large training set into a small synthetic set that preserves downstream training utility~\cite{wang2018dataset}. Existing methods typically optimize synthetic data by matching gradients~\cite{zhao2020dataset,n2,LD}, training trajectories~\cite{cazenavette2022dataset,cui2023scaling,guo2024datm,lee2024selmatch,zhong2025mct}, feature distributions~\cite{wang2022cafe,DM,DCC,point2,ran2026tgdd,li2025hdd,liu2025wasserstein}, generative processes~\cite{li2024generative,li2025generative,d4m,ye2025igds,chansantiago2026learnability}, synthetic-data parameterizations~\cite{kim2022dataset,deng2022remember,jiang2025gsdd,tran2026ptqdc}, or kernel-based objectives~\cite{KIP,nguyen2020dataset,zhou2022dataset,loo2022efficient}. These approaches have substantially advanced DD, but most of them are designed for training networks from scratch. As a result, they often require repeated inner-loop optimization, long-horizon meta-gradient computation, or strong differentiable augmentation. In contrast, we study DD in the frozen-backbone linear-probing regime, where the encoder is fixed, and the inner adaptation reduces to a much simpler convex problem with an exact closed-form solution.

\textbf{Dataset Distillation for Pre-trained Vision Models.}
The closest work to ours is Linear Gradient Matching (LGM)~\cite{cazenavette2025dataset}, which formulates DD for pre-trained self-supervised vision models. LGM optimizes synthetic images by matching gradients induced by a linear probe on frozen features, providing a strong benchmark for low-IPC distillation in this setting. However, it requires differentiation through the gradient-matching objective and achieves its best performance when combined with Differentiable Siamese Augmentation (DSA)~\cite{dsa}, which substantially increases the computational cost. CLP-DD addresses the same frozen-backbone setting from a complementary perspective. Instead of matching classifier gradients, it directly computes the linear classifier induced by the synthetic set through a closed-form linear-probe solver, and optimizes the synthetic images using a discriminative outer objective evaluated on real features.

\textbf{Implicit and Closed-form Optimization.}
Meta-learning and DD methods often obtain meta-gradients by differentiating through an inner optimization process, which becomes costly when the inner trajectory is long~\cite{zhou2022dataset}. Prior work has therefore explored implicit differentiation and equilibrium formulations to avoid explicit unrolling~\cite{rajeswaran2019meta,bai2019deep}. CLP-DD follows the same high-level goal of removing expensive inner-loop unrolling, but differs in a key respect: it does not approximate the solution of a non-convex training process. Since the only trainable component in our setting is a ridge-regularized linear classifier over fixed features, the inner problem admits an exact ridge-regression solution. In the low-IPC regime, this solution can be equivalently expressed in sample-space kernel form, allowing linear-probe adaptation to be implemented as a differentiable algebraic solve rather than an unrolled training trajectory.

\section{Method}
\label{sec:method}

We consider dataset distillation in the frozen-backbone linear-probing regime. Let $\phi(\cdot)$ denote a pre-trained feature extractor whose parameters are fixed throughout distillation. Given a synthetic image set $\mathcal{S}_{\mathrm{syn}}$ with $N$ samples, we denote its feature matrix by
\begin{equation}
X_{\mathrm{syn}} = \phi(\mathcal{S}_{\mathrm{syn}}) \in \mathbb{R}^{N \times d},
\end{equation}
where $d$ is the feature dimension. The corresponding one-hot label matrix is $Y_{\mathrm{syn}} \in \mathbb{R}^{N \times C}$, where $C$ is the number of classes. We use $\mathcal{S}_{\mathrm{real}}$ to denote real training samples used to define the outer distillation objective, and write $X_{\mathrm{real}}=\phi(\mathcal{S}_{\mathrm{real}})$. The downstream model is a linear probe $W \in \mathbb{R}^{d \times C}$ trained on top of the frozen features.

Our goal is to optimize $\mathcal{S}_{\mathrm{syn}}$ such that a linear classifier trained on its frozen features performs well on real data. Instead of differentiating through a sequence of linear-probe updates, we exploit the fact that the ridge-regularized linear-probe problem admits a closed-form solution. We refer to the resulting framework as Closed-Form Linear-Probe Dataset Distillation (CLP-DD). Detailed derivations corresponding to the following components are provided in Appendix~\ref{app:detailed_derivations}.

\subsection{Overview}
\label{subsec:method_overview}

CLP-DD separates closed-form inner adaptation from discriminative outer optimization. The inner problem models linear-probe adaptation on synthetic features with a ridge-regularized quadratic objective, which admits a closed-form solution. The outer problem then updates the synthetic images by evaluating the resulting classifier on real training samples. This separation keeps the inner adaptation analytically tractable while using real data to provide a discriminative optimization signal.

Concretely, each CLP-DD iteration consists of three steps. First, we extract frozen features for the current synthetic images. Second, we compute the linear-probe classifier induced by the synthetic set using a sample-space closed-form solver. Third, we evaluate an outer class-anchor loss on real features and backpropagate through the solver to update the synthetic images. Algorithm~\ref{alg:clpdd} summarizes the overall procedure.

\subsection{Closed-Form Linear-Probe Solver}
\label{subsec:closed_form_solver}

The inner optimization is defined on the synthetic features. We consider the ridge-regularized quadratic objective
\begin{equation}
\mathcal{L}_{\mathrm{in}}(W)
= \frac{1}{2}\|X_{\mathrm{syn}}W - Y_{\mathrm{syn}}\|_F^2
+ \frac{\lambda}{2}\|W\|_F^2,
\label{eq:inner_objective}
\end{equation}
where $\lambda > 0$ is the ridge coefficient. Since the encoder is fixed, the only trainable component in the inner problem is a linear classifier over frozen features. The objective is therefore strongly convex and admits the standard ridge-regression solution
\begin{equation}
W^*
=
(X_{\mathrm{syn}}^\top X_{\mathrm{syn}}+\lambda I_d)^{-1}
X_{\mathrm{syn}}^\top Y_{\mathrm{syn}}.
\label{eq:ridge_solution}
\end{equation}
Thus, linear-probe adaptation on the synthetic set can be computed directly, without differentiating through an unrolled sequence of classifier updates.

For completeness, Eq.~\eqref{eq:ridge_solution} can also be viewed as the steady state of gradient descent on Eq.~\eqref{eq:inner_objective}. Defining
\begin{equation}
H = X_{\mathrm{syn}}^\top X_{\mathrm{syn}} + \lambda I_d,
\qquad
c = X_{\mathrm{syn}}^\top Y_{\mathrm{syn}},
\label{eq:H_c_def}
\end{equation}
gradient descent with step size $\eta>0$ follows
\begin{equation}
W_{t+1}
= W_t-\eta(HW_t-c)
= (I-\eta H)W_t+\eta c.
\label{eq:gd_dynamics}
\end{equation}
Under the standard spectral-radius condition $\rho(I-\eta H)<1$, this recursion converges to the unique fixed point satisfying $HW^*=c$, which recovers Eq.~\eqref{eq:ridge_solution}. We provide the derivation in  Appendix~\ref{app:steady_state}.

\subsection{Sample-Space Kernel Reformulation}
\label{subsec:kernel_reformulation}

Directly computing Eq.~\eqref{eq:ridge_solution} requires solving a $d\times d$ linear system. In the low-IPC regime, the number of synthetic samples is much smaller than the feature dimension, i.e., $N\ll d$. We therefore use the equivalent sample-space kernel form
\begin{equation}
W^*
=
X_{\mathrm{syn}}^\top
(X_{\mathrm{syn}}X_{\mathrm{syn}}^\top+\lambda I_N)^{-1}
Y_{\mathrm{syn}}.
\label{eq:kernel_form}
\end{equation}
The equivalence follows from the standard push-through identity:
\begin{equation}
(X_{\mathrm{syn}}^\top X_{\mathrm{syn}}+\lambda I_d)^{-1}X_{\mathrm{syn}}^\top
=
X_{\mathrm{syn}}^\top
(X_{\mathrm{syn}}X_{\mathrm{syn}}^\top+\lambda I_N)^{-1},
\label{eq:push_through}
\end{equation}
where $\lambda>0$ ensures that the regularized systems are positive definite and invertible. Let
\begin{equation}
K_{\mathrm{syn}}=X_{\mathrm{syn}}X_{\mathrm{syn}}^\top \in \mathbb{R}^{N\times N}.
\label{eq:kernel_matrix}
\end{equation}
Then the solver only needs to form the synthetic sample kernel and solve an $N\times N$ linear system, where $N=C\times\mathrm{IPC}$. This shifts the computational bottleneck from the feature dimension to the number of synthetic samples, making the formulation well-suited to extreme low-IPC distillation. Importantly, this closed-form step removes the dependence on an unrolled classifier-training horizon, although gradients still need to be backpropagated through the frozen encoder to update the synthetic pixels. When $N$ is not smaller than $d$, the primal form in Eq.~\eqref{eq:ridge_solution} can be used instead without changing the induced classifier. We provide the derivation of the equivalence in Appendix~\ref{app:kernel_reformulation}.

\begin{algorithm}[t]
\caption{Closed-Form Linear-Probe Dataset Distillation (CLP-DD)}
\label{alg:clpdd}
\begin{algorithmic}[1]
\REQUIRE Frozen feature extractor $\phi$; synthetic images $\mathcal{S}_{\mathrm{syn}}$ and labels $Y_{\mathrm{syn}}$; real training samples $\mathcal{S}_{\mathrm{real}}$; ridge coefficient $\lambda$; temperature $\tau$
\ENSURE Optimized synthetic set $\mathcal{S}_{\mathrm{syn}}$
\FOR{$t=1,\ldots,4{,}000$}
    \STATE Extract synthetic features:
    $X_{\mathrm{syn}} \gets \phi(\mathcal{S}_{\mathrm{syn}})$
    \STATE Sample a class-balanced real batch from $\mathcal{S}_{\mathrm{real}}$ and extract features:
    $X_{\mathrm{real}} \gets \phi(\mathcal{S}_{\mathrm{real}})$
    \STATE Form the synthetic kernel matrix:
    $K_{\mathrm{syn}} \gets X_{\mathrm{syn}}X_{\mathrm{syn}}^{\top}$
    \STATE Solve the closed-form linear probe:
    $W^* \gets X_{\mathrm{syn}}^{\top}(K_{\mathrm{syn}}+\lambda I_N)^{-1}Y_{\mathrm{syn}}$
    \STATE Compute the outer loss $\mathcal{L}_{\mathrm{meta}}$ on $X_{\mathrm{real}}$ using Eq.~\eqref{eq:meta_loss}
    \STATE Backpropagate through the closed-form solve to obtain $\nabla_{\mathcal{S}_{\mathrm{syn}}}\mathcal{L}_{\mathrm{meta}}$
    \STATE Update the synthetic images $\mathcal{S}_{\mathrm{syn}}$
\ENDFOR
\RETURN $\mathcal{S}_{\mathrm{syn}}$
\end{algorithmic}
\end{algorithm}

\subsection{Outer Class-Anchor Objective}
\label{subsec:outer_objective}

The closed-form solver specifies how a linear probe is induced by the current synthetic features. To optimize the synthetic images, we further evaluate this induced classifier on real training features and use the resulting discriminative signal as the outer objective.

Given $W^*$, the outer loss is
\begin{equation}
\mathcal{L}_{\mathrm{meta}}
=
-\frac{1}{M}\sum_{i=1}^{M}
\log
\frac{
\exp(x_{\mathrm{real},i}^{\top}w^*_{y_i}/\tau)
}{
\sum_{j=1}^{C}
\exp(x_{\mathrm{real},i}^{\top}w^*_{j}/\tau)
},
\label{eq:meta_loss}
\end{equation}
where $M$ is the number of real samples in the outer batch, i.e., $M=CB$ for a class-balanced batch with $B$ real samples per class, $\tau$ is a temperature parameter, and $w_j^*$ denotes the $j$-th column of $W^*$.

We refer to Eq.~\eqref{eq:meta_loss} as the class-anchor cross-entropy objective. It treats the columns of the closed-form classifier as class anchors in the frozen feature space and encourages real samples to score highly against the anchor of their ground-truth class while being separated from competing class anchors. Compared with a squared loss on one-hot targets, this objective directly optimizes the softmax decision rule used by the final linear probe, and therefore provides a discriminative outer signal more aligned with transfer accuracy. The gradient of this objective with respect to the closed-form classifier is derived in Appendix~\ref{app:outer_loss_gradient}.

\subsection{Differentiating Through the Solver}
\label{subsec:implicit_diff}

CLP-DD updates the synthetic images by differentiating the outer loss through the closed-form solver. From Eq.~\eqref{eq:kernel_form}, let
\begin{equation}
A = K_{\mathrm{syn}}+\lambda I_N.
\label{eq:A_def}
\end{equation}
Then
\begin{equation}
W^*
=
X_{\mathrm{syn}}^\top A^{-1}Y_{\mathrm{syn}}.
\label{eq:wstar_kernel}
\end{equation}
The gradient of $\mathcal{L}_{\mathrm{meta}}$ with respect to $X_{\mathrm{syn}}$ is obtained by differentiating through both $X_{\mathrm{syn}}^\top$ and $A^{-1}$. The derivative of the inverse follows
\begin{equation}
\mathrm{d}(A^{-1})
=
-A^{-1}\mathrm{d}A A^{-1}.
\label{eq:inverse_derivative}
\end{equation}
Since $A$ depends on $X_{\mathrm{syn}}$ through $K_{\mathrm{syn}}=X_{\mathrm{syn}}X_{\mathrm{syn}}^\top$, gradients from the outer loss propagate to the synthetic features and then to the synthetic images through the frozen feature extractor. In practice, this differentiation is implemented with automatic differentiation through the linear solve, avoiding the need to store or backpropagate through an unrolled inner trajectory. Detailed derivations for differentiating through the closed-form solver and propagating the gradient to synthetic images are provided in Appendices~\ref{app:kernel_gradient} and~\ref{app:synthetic_image_gradient}.

% Keep Tables 1 and 2 on separate pages, with prose below each table.
\setcounter{topnumber}{1}
\begin{table}[!t]
\centering
\small
\caption{Quantitative performance comparison on ImageNet-100 and ImageNet-1K at IPC$=$1. We report the mean and standard deviation over three repeated runs. Bold indicates the best DSA-free low-IPC result on ImageNet-100 and the best low-IPC result on ImageNet-1K; full-data results are references.}
\label{tab:baseline_comparison}
\begin{tabular}{ll cccc c} 
\toprule
\multirow{2}{*}{\textbf{Dataset}} & \multirow{2}{*}{\textbf{Method}} & \multicolumn{4}{c}{\textbf{Backbone Architectures}} & \multirow{2}{*}{\textbf{Average}} \\
\cmidrule(lr){3-6} 
& & \textbf{DINOv2} & \textbf{MoCo-v3} & \textbf{CLIP} & \textbf{EVA-02} & \\
\midrule
\multirow{9}{*}{ImageNet-100} 
& Neighbor    & 86.0$\textcolor{lightgray}{\pm 0.2}$ & 77.1$\textcolor{lightgray}{\pm 0.2}$ & 67.8$\textcolor{lightgray}{\pm 0.2}$ & 78.8$\textcolor{lightgray}{\pm 0.1}$ & 77.4$\textcolor{lightgray}{\pm 0.2}$ \\
& Centroid    & 86.9$\textcolor{lightgray}{\pm 0.3}$ & 77.7$\textcolor{lightgray}{\pm 0.1}$ & 77.1$\textcolor{lightgray}{\pm 0.1}$ & 80.9$\textcolor{lightgray}{\pm 0.2}$ & 80.6$\textcolor{lightgray}{\pm 0.2}$ \\
& Random      & 74.8$\textcolor{lightgray}{\pm 2.6}$ & 61.3$\textcolor{lightgray}{\pm 2.6}$ & 57.6$\textcolor{lightgray}{\pm 1.5}$ & 64.4$\textcolor{lightgray}{\pm 2.5}$ & 64.5$\textcolor{lightgray}{\pm 2.5}$ \\
& k-center    & \textbf{90.2$\textcolor{lightgray}{\pm 0.2}$} & 78.6$\textcolor{lightgray}{\pm 0.6}$ & 75.8$\textcolor{lightgray}{\pm 0.2}$ & 81.1$\textcolor{lightgray}{\pm 0.1}$ & 81.4$\textcolor{lightgray}{\pm 0.3}$ \\
& LFP         & 88.4$\textcolor{lightgray}{\pm 0.2}$ & 81.7$\textcolor{lightgray}{\pm 0.1}$ & 78.3$\textcolor{lightgray}{\pm 0.3}$ & 85.3$\textcolor{lightgray}{\pm 0.8}$ & 83.4$\textcolor{lightgray}{\pm 0.4}$ \\
& LGM         & 82.6$\textcolor{lightgray}{\pm 0.2}$ & 59.5$\textcolor{lightgray}{\pm 0.5}$ & 58.4$\textcolor{lightgray}{\pm 0.3}$ & 73.9$\textcolor{lightgray}{\pm 0.2}$ & 68.6$\textcolor{lightgray}{\pm 0.4}$ \\
& \textbf{CLP-DD (Ours)} & 89.2$\textcolor{lightgray}{\pm 0.1}$ & \textbf{83.9$\textcolor{lightgray}{\pm 0.1}$} & \textbf{79.2$\textcolor{lightgray}{\pm 0.1}$} & \textbf{87.6$\textcolor{lightgray}{\pm 0.0}$} & \textbf{85.0$\textcolor{lightgray}{\pm 0.1}$} \\
\cmidrule(lr){2-7}
& \textcolor{gray}{LGM w/ DSA} & \textcolor{gray}{91.4}$\textcolor{lightgray}{\pm 0.1}$ & \textcolor{gray}{83.3}$\textcolor{lightgray}{\pm 0.1}$ & \textcolor{gray}{84.6}$\textcolor{lightgray}{\pm 0.1}$ & \textcolor{gray}{88.6}$\textcolor{lightgray}{\pm 0.1}$ & \textcolor{gray}{87.0}$\textcolor{lightgray}{\pm 0.1}$ \\
& Full dataset & 95.1$\textcolor{lightgray}{\pm 0.1}$ & 89.3$\textcolor{lightgray}{\pm 0.2}$ & 92.4$\textcolor{lightgray}{\pm 0.0}$ & 94.0$\textcolor{lightgray}{\pm 0.1}$ & 92.7$\textcolor{lightgray}{\pm 0.1}$ \\
\midrule
\multirow{6}{*}{ImageNet-1K} 
& Neighbor    & 67.7$\textcolor{lightgray}{\pm 0.1}$ & 56.3$\textcolor{lightgray}{\pm 0.1}$ & 38.7$\textcolor{lightgray}{\pm 0.1}$ & 50.0$\textcolor{lightgray}{\pm 0.1}$ & 53.2$\textcolor{lightgray}{\pm 0.1}$ \\
& Centroid    & 69.5$\textcolor{lightgray}{\pm 0.0}$ & 57.4$\textcolor{lightgray}{\pm 0.1}$ & 53.8$\textcolor{lightgray}{\pm 0.1}$ & 58.0$\textcolor{lightgray}{\pm 0.0}$ & 59.6$\textcolor{lightgray}{\pm 0.1}$ \\
& Random      & 50.3$\textcolor{lightgray}{\pm 0.5}$ & 38.7$\textcolor{lightgray}{\pm 0.5}$ & 31.7$\textcolor{lightgray}{\pm 0.4}$ & 37.7$\textcolor{lightgray}{\pm 0.6}$ & 39.6$\textcolor{lightgray}{\pm 0.5}$ \\
& LGM w/ DSA  & 75.0$\textcolor{lightgray}{\pm 0.1}$ & 63.2$\textcolor{lightgray}{\pm 0.0}$ & \textbf{62.9$\textcolor{lightgray}{\pm 0.0}$} & 70.3$\textcolor{lightgray}{\pm 0.1}$ & 67.8$\textcolor{lightgray}{\pm 0.1}$ \\
& \textbf{CLP-DD (Ours)} & \textbf{75.6$\textcolor{lightgray}{\pm 0.1}$} & \textbf{63.4$\textcolor{lightgray}{\pm 0.1}$} & 62.4$\textcolor{lightgray}{\pm 0.1}$ & \textbf{71.2$\textcolor{lightgray}{\pm 0.1}$} & \textbf{68.2$\textcolor{lightgray}{\pm 0.1}$} \\
\cmidrule(lr){2-7}
& Full dataset & 83.0$\textcolor{lightgray}{\pm 0.1}$ & 76.5$\textcolor{lightgray}{\pm 0.0}$ & 78.7$\textcolor{lightgray}{\pm 0.0}$ & 81.7$\textcolor{lightgray}{\pm 0.0}$ & 80.0$\textcolor{lightgray}{\pm 0.1}$ \\
\bottomrule
\end{tabular}
\end{table}

\section{Experiments}
\label{sec:experiments}

We evaluate \textbf{CLP-DD} on ImageNet-100~\cite{tian2020contrastive} and ImageNet-1K~\cite{russakovsky2015imagenet}. We use four representative pre-trained vision backbones: CLIP~\cite{radford2021learning}, DINOv2~\cite{oquab2024dinov2}, EVA-02~\cite{fang2023eva}, and MoCo-v3~\cite{chen2021empirical}. Unless otherwise specified, all experiments are conducted at a resolution of $224\times224$ using the ViT-B variant of each backbone. We distill each dataset for a fixed $4,000$ iterations, without validation- or test-based checkpoint selection, and evaluate the final distilled images by training a randomly initialized linear classifier to convergence on top of the frozen features. Accuracy results are reported as mean $\pm$ standard deviation over three independent runs unless indicated otherwise.

At each distillation iteration, we sample a class-balanced real batch with $B=4$ samples per class to define the outer objective. We analyze the effect of $B$ in the ablation study. Unless otherwise specified, we set the ridge coefficient to $\lambda=0.1$ and the temperature to $\tau=0.07$, and analyze their sensitivity in the ablation study. 

\begin{wrapfigure}{r}{0.4\textwidth}
    \vspace{-4mm}
    \centering
    \includegraphics[width=\linewidth]{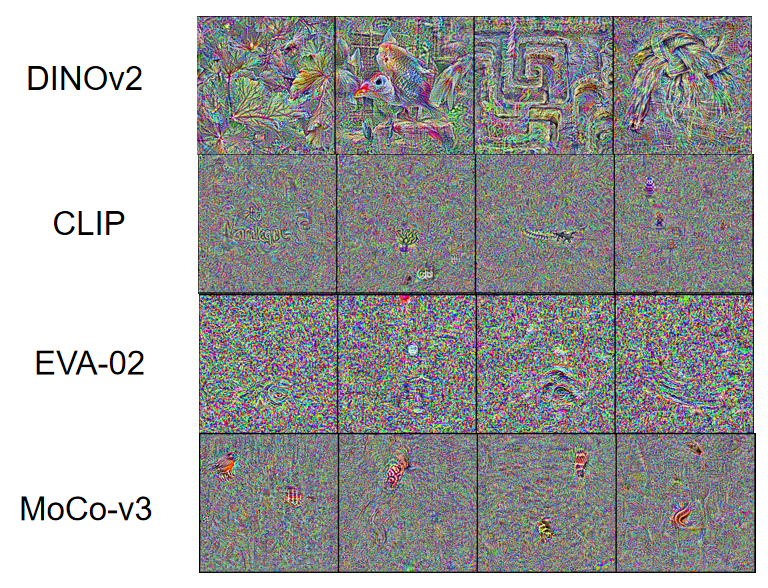}
    \caption{Examples of IPC$=$1 images distilled by CLP-DD on ImageNet-100 using the CLIP, DINOv2, MoCo-v3 and EVA-02 backbone.}
    \label{fig:CLP-DD_on_4_examples}
    \vspace{-4mm}
\end{wrapfigure}

By default, CLP-DD uses standard image augmentations, including random cropping and noise injection, and does not use Differentiable Siamese Augmentation (DSA)~\cite{dsa}. For controlled evaluation, we extract features from the final layer of each backbone and use the same linear-probe training protocol across all methods and baselines. Additional implementation details are provided in Appendix~\ref{app:implementation_details}.

Figure~\ref{fig:CLP-DD_on_4_examples} shows examples of images distilled by CLP-DD with the CLIP, DINOv2, MoCo-v3, and EVA-02 backbones. Rows correspond to different backbones, and columns correspond to
the first four ImageNet-100 categories. As is common in pixel-space dataset distillation, the distilled images are highly textured and do not resemble natural images. This is expected because CLP-DD optimizes synthetic images for frozen-feature linear-probe performance rather than visual realism. We therefore use downstream linear-probe accuracy as the primary evaluation metric. Additional distilled images are provided in Appendix~\ref{app:additional_distilled_images}.

\subsection{Standard ImageNet Evaluation}
\label{subsec:main_evaluation}

We evaluate each distilled set by training a linear classifier on the synthetic images and reporting accuracy on the corresponding ImageNet validation split, using the same protocol for all real-image baselines. We compare with three IPC$=$1 real-image baselines: Random selects one image per class uniformly at random; Centroid selects the image closest to the class-wise mean feature; and Neighbor selects the real image closest to the corresponding CLP-DD synthetic image in feature space. On ImageNet-100, we additionally compare with class-wise k-center selection and Learned Feature Prototypes (LFP). For LGM, we report both the DSA-free and DSA-based variants when available.

Table~\ref{tab:baseline_comparison} compares CLP-DD with real-image baselines and LGM on ImageNet-100 and ImageNet-1K at IPC$=$1. CLP-DD consistently outperforms Random, Centroid, and Neighbor across all four backbones on both datasets, improving the four-backbone average over k-center by 3.6 points and over LFP by 1.6 points on ImageNet-100, and over the strongest real-image baseline by 8.6 percentage points on ImageNet-1K. k-center is stronger on DINOv2 (90.2\% versus 89.2\%), so the ImageNet-100 advantage is not uniform across backbones. Compared with LGM, CLP-DD substantially improves over the DSA-free variant on ImageNet-100, from 68.6\% to 85.0\%. Compared with LGM w/ DSA, CLP-DD trails by 2.0 points on ImageNet-100 (85.0\% vs.\ 87.0\%) and is comparable on ImageNet-1K (68.2\% versus 67.8\%); the small per-backbone differences are interpreted descriptively. These results position CLP-DD as an efficient DSA-free alternative to linear gradient matching.

\newcommand{\std}[1]{{\color{gray}$\pm$#1}}
\begin{table*}[t]
\centering
\caption{Quantitative performance comparison on Spawrious and WaterBirds at IPC$=$1. We report the mean and standard deviation over three repeated runs. Bold indicates the best DSA-free low-IPC result; gray rows are DSA-based references.}
\label{tab:combined_results_v2}
\resizebox{0.95\textwidth}{!}{
\begin{tabular}{ll ccccc}
\toprule
\multirow{2}{*}{\textbf{Dataset}} & \multirow{2}{*}{\textbf{Method}} & \multicolumn{4}{c}{\textbf{Backbone Architectures}} & \multirow{2}{*}{\textbf{Average}} \\
\cmidrule(lr){3-6}
& & \textbf{DINOv2} & \textbf{MoCo-v3} & \textbf{CLIP} & \textbf{EVA-02} & \\
\midrule
\multirow{6}{*}{Spawrious}
& Neighbor & 76.9\std{1.2} & 33.0\std{3.4} & 41.7\std{4.6} & \textbf{40.7}\std{2.7} & 48.1\std{3.0} \\
& Centroid & \textbf{80.2}\std{2.7} & 30.5\std{1.6} & 44.8\std{5.0} & 38.1\std{3.3} & \textbf{48.4}\std{3.2} \\
& Random   & 68.1\std{8.1} & 31.8\std{2.5} & \textbf{46.2}\std{4.4} & 33.3\std{5.3} & 44.9\std{5.1} \\
& LGM      & 57.2\std{6.8} & 22.6\std{4.3} & 26.1\std{6.7} & 28.7\std{2.1} & 33.6\std{5.0} \\
& \textbf{CLP-DD (Ours)} & 78.3\std{2.2}    & \textbf{33.2}\std{2.7}    & 42.7\std{4.8}    & 37.4\std{3.5}    & 47.9\std{3.3} \\
\cmidrule(lr){2-7}
& \textcolor{gray}{LGM w/ DSA}   & \textcolor{gray}{80.8}\std{3.1} & \textcolor{gray}{32.7}\std{2.3} & \textcolor{gray}{43.1}\std{6.4} & \textcolor{gray}{36.5}\std{3.1} & \textcolor{gray}{48.3}\std{3.7} \\
& {Full dataset} & {78.1}\std{0.1} & {35.8}\std{0.0} & {50.4}\std{0.3}  & {50.1}\std{0.4} & {53.6}\std{0.2} \\
\midrule
\multirow{6}{*}{WaterBirds}
& Neighbor & 67.3\std{5.8} & 45.2\std{11.4} & 74.6\std{6.5} & 74.3\std{3.3} & 65.3\std{6.8} \\
& Centroid & 65.7\std{3.4} & 62.0\std{5.2} & 69.8\std{8.6} & 64.6\std{7.9} & 65.5\std{6.3} \\
& Random   & 57.1\std{8.6} & 67.4\std{5.7} & 71.9\std{4.5} & 59.8\std{13.3} & 64.0\std{8.0} \\
& LGM      & 58.3\std{3.2} & 63.4\std{1.6} & 64.1\std{2.2} & 64.8\std{1.8} & 62.6\std{2.2} \\
& \textbf{CLP-DD (Ours)} & \textbf{80.0}\std{2.3}    & \textbf{78.5}\std{0.4}    & \textbf{75.4}\std{0.3}    & \textbf{79.6}\std{0.4}    & \textbf{78.3}\std{0.8} \\
\cmidrule(lr){2-7}
& \textcolor{gray}{LGM w/ DSA}     & \textcolor{gray}{82.1}\std{2.9} & \textcolor{gray}{77.8}\std{0.0} & \textcolor{gray}{77.9}\std{0.2} & \textcolor{gray}{78.0}\std{0.8} & \textcolor{gray}{79.0}\std{1.0} \\
& {Full dataset} & {95.5}\std{0.1} & {74.3}\std{0.1} & {86.0}\std{0.1} & {90.4}\std{0.2} & {86.5}\std{0.1} \\
\bottomrule
\end{tabular}
}
\end{table*}

\subsection{Auxiliary Evaluation under Background Shift}
\label{subsec:spurious_evaluation}

We further evaluate CLP-DD on Spawrious and WaterBirds, following the protocol of LGM. Unlike standard ImageNet evaluation, these benchmarks test performance under background shift: class labels are spuriously correlated with background attributes during training, while evaluation is conducted under shifted or adversarial background configurations.

Table~\ref{tab:combined_results_v2} summarizes the results. On Spawrious~\cite{hayes2023spawrious}, CLP-DD achieves 47.9\% average accuracy, competitive with Centroid (48.4\%), Neighbor (48.1\%), and LGM w/ DSA (48.3\%), while substantially outperforming DSA-free LGM (33.6\%). All low-IPC methods, including the strongest real-image baselines, remain below the full-dataset upper bound of 53.6\%, indicating that severe background shift remains challenging in this regime. 
On WaterBirds~\cite{sagawa2019distributionally}, CLP-DD obtains 78.3\% average accuracy, close to LGM w/ DSA at 79.0\%, and even achieves better results on MoCo-v3 and EVA-02 compared to LGM w/ DSA.

\subsection{IPC Scaling}

\begin{wraptable}{r}{0.48\textwidth}
    \vspace{-4mm}
    \centering
    \small
    \setlength{\tabcolsep}{4pt}
    \renewcommand{\arraystretch}{0.95}
    \caption{IPC scaling on ImageNet-100 with DINOv2.}
    \label{tab:dinov2_in100}
    \begin{tabular}{lccc}
        \toprule
        \multirow{2}{*}{\textbf{Method}} & \multicolumn{3}{c}{\textbf{IPC}} \\
        \cmidrule(lr){2-4}
        & \textbf{1} & \textbf{3} & \textbf{5} \\
        \midrule
        Centroid    & 86.9$\textcolor{lightgray}{\pm 0.3}$ & 89.3$\textcolor{lightgray}{\pm 0.2}$ & 90.2$\textcolor{lightgray}{\pm 0.2}$ \\
        % LGM w/ DSA  & 91.4$\textcolor{lightgray}{\pm 0.1}$ & 92.6$\textcolor{lightgray}{\pm 0.0}$ & 92.9$\textcolor{lightgray}{\pm 0.1}$ \\
        LGM         & 82.6$\textcolor{lightgray}{\pm 0.2}$ & 86.3$\textcolor{lightgray}{\pm 0.2}$ & 87.8$\textcolor{lightgray}{\pm 0.3}$ \\
        \textbf{CLP-DD} & \textbf{89.2$\textcolor{lightgray}{\pm 0.1}$} & \textbf{91.3$\textcolor{lightgray}{\pm 0.1}$} & \textbf{91.8$\textcolor{lightgray}{\pm 0.0}$} \\
        \bottomrule
    \end{tabular}
    \vspace{-4mm}
\end{wraptable}

Table~\ref{tab:dinov2_in100} reports IPC scaling on ImageNet-100 with DINOv2. CLP-DD outperforms both Centroid and LGM at each tested IPC. The gains over LGM are 6.6, 5.0, and 4.0 percentage points at IPC$=1$, $3$, and $5$, respectively, without DSA. Appendix~\ref{app:ipc_scaling} reports additional results.

\WFclear
\subsection{Efficiency Analysis}

\begin{wraptable}{r}{0.45\textwidth}
    \vspace{-4mm}
    \centering
    \small
    \setlength{\tabcolsep}{3.5pt}
    \renewcommand{\arraystretch}{0.95}
    \caption{ImageNet-100/DINOv2 efficiency at IPC$=$1. Time is seconds per 100 full iterations; memory is peak MB.}
    \label{tab:memory_comparison}
    \begin{tabular}{lccc}
        \toprule
        Method & Time & Memory & Acc. \\
        \midrule
        LGM w/ DSA & 57.49s & 19,320MB & 91.4$\textcolor{lightgray}{\pm 0.1}$ \\
        LGM & 4.25s & 2,334MB & 82.6$\textcolor{lightgray}{\pm 0.2}$ \\
        \textbf{CLP-DD} & 4.07s & 2,361MB & 89.2$\textcolor{lightgray}{\pm 0.1}$ \\
        \bottomrule
    \end{tabular}
    \vspace{-4mm}
\end{wraptable}

Table~\ref{tab:memory_comparison} compares the empirical efficiency of CLP-DD against LGM baselines on a single NVIDIA RTX 5090 GPU. All methods are evaluated with the same DINOv2 backbone and image resolution. CLP-DD has a similar time and memory footprint to LGM without DSA, while achieving substantially higher accuracy. In contrast, LGM with DSA obtains the highest accuracy in this setting, but requires substantially more memory and longer runtime due to the additional augmentation pipeline. Time measures 100 complete distillation iterations, including feature extraction, augmentation, loss computation, backward propagation, image updates, and the differentiable solve where applicable. One-time setup and final evaluation are excluded for all methods. CLP-DD approaches the accuracy of LGM with DSA at a cost close to DSA-free LGM.

\WFclear
\begin{wrapfigure}{r}{0.3\textwidth}
    \vspace{-4mm}
    \centering
    \includegraphics[width=\linewidth]{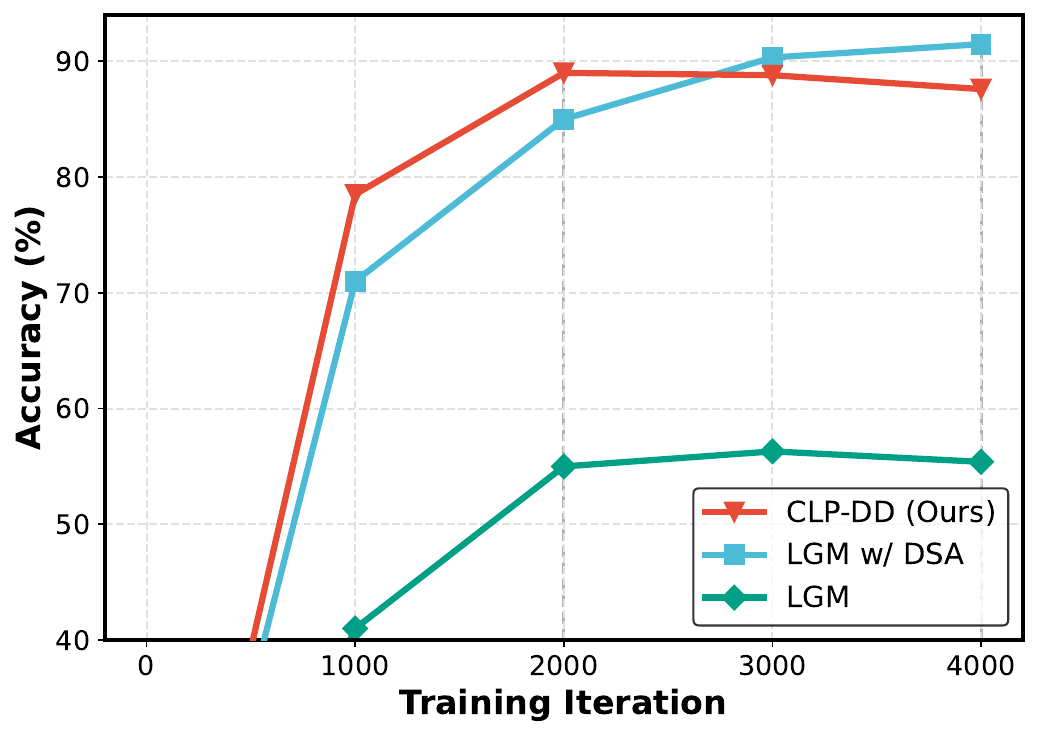}
    \caption{Convergence comparison on ImageNet-100 with DINOv2 at IPC$=$1.}
    \label{fig:CLP-DD_LGM_dino_Com}
    \vspace{-4mm}
\end{wrapfigure}

Figure~\ref{fig:CLP-DD_LGM_dino_Com} compares the convergence behavior of CLP-DD with LGM variants under the same distillation budget. CLP-DD reaches high accuracy in fewer iterations than DSA-free LGM and maintains a clear advantage throughout most of the optimization process. LGM w/ DSA also achieves strong final accuracy, but as shown in Table~\ref{tab:memory_comparison}, this comes with substantially higher runtime and memory cost. These results suggest that the closed-form linear-probe solver provides a more direct optimization signal than linear gradient matching in the DSA-free setting. The mild late-stage accuracy decline occurs during outer image optimization, since the inner ridge solution is exact. Its mechanism remains unresolved; training-only stopping criteria or additional regularization are future work.

\begingroup
\setlength{\intextsep}{4pt}
\subsection{Ablation Study}

\begin{table}[H]
\centering
\small
\caption{Ablation of the outer objective on ImageNet-100 at IPC$=$1. ``w/o CA'' replaces the class-anchor objective with an MSE objective.}
\label{tab:ablation_CLP-DD}
\begin{tabular}{lccccc}
\toprule
\textbf{Method} & \textbf{DINOv2} & \textbf{MoCo-v3} & \textbf{CLIP} & \textbf{EVA-02} & \textbf{Average} \\
\midrule
LGM         & 82.6$\textcolor{lightgray}{\pm 0.2}$ & 59.5$\textcolor{lightgray}{\pm 0.5}$ & 58.4$\textcolor{lightgray}{\pm 0.3}$ & 73.9$\textcolor{lightgray}{\pm 0.2}$ & 68.6$\textcolor{lightgray}{\pm 0.4}$ \\
CLP-DD w/o CA 
& 88.3$\textcolor{lightgray}{\pm 0.0}$ 
& 82.1$\textcolor{lightgray}{\pm 0.1}$ 
& 74.1$\textcolor{lightgray}{\pm 0.1}$ 
& 84.1$\textcolor{lightgray}{\pm 0.1}$ 
& 82.1$\textcolor{lightgray}{\pm 0.1}$ \\
\textbf{CLP-DD (Ours)} 
& \textbf{89.2$\textcolor{lightgray}{\pm 0.1}$} 
& \textbf{83.9$\textcolor{lightgray}{\pm 0.1}$} 
& \textbf{79.2$\textcolor{lightgray}{\pm 0.1}$} 
& \textbf{87.6$\textcolor{lightgray}{\pm 0.0}$} 
& \textbf{85.0$\textcolor{lightgray}{\pm 0.1}$} \\
\bottomrule
\end{tabular}
\end{table}

\textbf{Effectiveness of the Class-Anchor Outer Objective.}
We ablate the outer objective on ImageNet-100 at IPC$=$1 by replacing the temperature-scaled class-anchor objective with an ordinary MSE objective. As shown in Table~\ref{tab:ablation_CLP-DD}, the class-anchor objective improves the average accuracy from 82.1\% to 85.0\%, with consistent gains across all four backbones. The largest improvement is observed on CLIP, where the accuracy increases from 74.1\% to 79.2\%.

These results show that the outer objective contributes beyond the closed-form inner solver. While the sample-space solver computes the linear probe induced by the current synthetic set, the outer objective determines how the synthetic images are updated so that the induced classifier transfers to real samples. Compared with ordinary MSE, the class-anchor objective directly optimizes relative class discrimination in the frozen feature space, which appears more suitable for linear-probe dataset distillation.

\begin{wraptable}{r}{0.32\textwidth}
    \centering
    % \vspace{-4mm}
    \small
    \caption{Ablation on real samples per class ($B$).}
    \label{tab:real_samples_ablation}
    \begin{tabular}{cc}
        \toprule
        \textbf{$B$} & \textbf{Acc (\%)} \\
        \midrule
        1 & 88.0$\textcolor{lightgray}{\pm 0.1}$ \\
        2 & 88.9$\textcolor{lightgray}{\pm 0.2}$ \\
        \textbf{4} & \textbf{89.2$\textcolor{lightgray}{\pm 0.1}$} \\
        8 & 89.3$\textcolor{lightgray}{\pm 0.0}$ \\
        \bottomrule
    \end{tabular}
\end{wraptable}

\textbf{Effect of Real Batch Size ($B$).}
Table~\ref{tab:real_samples_ablation} varies the number $B$ of real samples per class on ImageNet-100 with DINOv2. Accuracy is 88.0\%, 88.9\%, 89.2\%, and 89.3\% at $B=1$, $2$, $4$, and $8$, respectively.

Each outer batch contains $CB$ real samples across $C$ classes,
with the synthetic set size fixed throughout the sweep. Successive
doublings of $B$ improve accuracy by 0.9, 0.3, and 0.1 percentage
points, respectively. We therefore use $B=4$ by default, retaining
most of the observed gain while using half as many real samples
per iteration as $B=8$.

\WFclear
\begin{wraptable}{r}{0.4\textwidth}
    \vspace{-4mm}
    \centering
    \small
    \caption{Sensitivity analysis of $\tau$ and $\lambda$ on ImageNet-100 with MoCo-v3 at IPC$=$1.}
    \label{tab:hyperparameters}
    \begin{tabular}{cc|cc}
        \toprule
        \textbf{$\tau$} & \textbf{Acc (\%)} & \textbf{$\lambda$} & \textbf{Acc (\%)} \\
        \midrule
        0.01 & 69.7$\textcolor{lightgray}{\pm 0.3}$ & 0.05 & 79.3$\textcolor{lightgray}{\pm 0.2}$ \\
        0.05 & 78.1$\textcolor{lightgray}{\pm 0.2}$ & 0.07 & 81.3$\textcolor{lightgray}{\pm 0.2}$ \\
        \textbf{0.07} & \textbf{83.9$\textcolor{lightgray}{\pm 0.1}$} & \textbf{0.10} & \textbf{83.9$\textcolor{lightgray}{\pm 0.1}$} \\
        0.10 & 83.2$\textcolor{lightgray}{\pm 0.0}$ & 0.50 & 76.2$\textcolor{lightgray}{\pm 0.1}$ \\
        \bottomrule
    \end{tabular}
\end{wraptable}

\textbf{Sensitivity Analysis on Hyperparameters.}
We analyze the sensitivity of CLP-DD to the temperature $\tau$ and the ridge coefficient $\lambda$ on ImageNet-100 with the MoCo-v3 backbone at IPC$=1$. We vary one hyperparameter at a time while keeping the other fixed to its default value. 

As shown in Table~\ref{tab:hyperparameters}, CLP-DD achieves the best accuracy among the tested temperatures at $\tau=0.07$. A smaller temperature, such as $\tau=0.01$, makes the softmax distribution overly sharp and leads to a noticeable accuracy drop, while $\tau=0.10$ remains competitive but slightly worse.
For the ridge coefficient, $\lambda=0.10$ yields the best result. Smaller values provide weaker regularization and may amplify noise in the induced classifier, whereas larger values over-regularize $W^*$ and reduce its discriminative capacity. Overall, CLP-DD remains stable across a moderate range of $\tau$ and $\lambda$, with performance degradation mainly observed under overly sharp temperatures or excessive ridge regularization.

The sweeps in Appendix~\ref{app:hyperparameter_sensitivity} show
that $\tau=0.07$ and $\lambda=0.10$ also give the highest mean
accuracy among the tested values on DINOv2, CLIP, and EVA-02.
Increasing $\lambda$ from 0.10 to 0.20 lowers accuracy by
1.6--2.6 percentage points across the four backbones.

\begin{wraptable}{r}{0.40\textwidth}
    \vspace{-3mm}
    \centering
    \small
    \setlength{\tabcolsep}{4pt}
    \caption{ImageNet-100/DINOv2 augmentation ablation at IPC$=$1. Mean $\pm$ std., three runs.}
    \label{tab:augmentation_main}
    \begin{tabular}{lc}
        \toprule
        CLP-DD setting & Acc. (\%) \\
        \midrule
        No augmentation & $87.9\pm0.8$ \\
        Cropping w/ noise & $89.2\pm0.1$ \\
        \bottomrule
    \end{tabular}
\end{wraptable}

\textbf{Effect of Standard Augmentation.}
Cropping and noise improve accuracy from 87.9\% to 89.2\%, a gain of 1.3 percentage points (Table~\ref{tab:augmentation_main}). Without augmentation, CLP-DD still exceeds DSA-free LGM (82.6\%) by 5.3 points, indicating that augmentation alone cannot explain the improvement. Standard augmentation therefore complements the proposed formulation, while a substantial advantage remains when it is removed. Nevertheless, this ablation does not fully disentangle the contributions of the closed-form inner learner, class-anchor outer objective, and augmentation to the gap between CLP-DD and LGM. Further augmentation and stability results are provided in Appendix~\ref{app:additional_experiments}.

Across three runs, cropping and noise also reduce the observed standard
deviation from 0.8 to 0.1 percentage points. Adding DSA ($m=1$) to the
default augmentation yields 89.3\% accuracy, with no reported standard
deviation (Table~\ref{tab:augmentation_ablation}). Relative to the default,
this 0.1-point difference comes with an increase from 4.07\,s to 9.39\,s
per 100 distillation iterations and from 2,361\,MB to 4,011\,MB in peak
GPU memory.

\WFclear
\endgroup
\begingroup
\setlength{\intextsep}{4pt}
\subsection{Cross-Backbone Transfer and ImageNet-1K Efficiency}
\label{subsec:transfer_scale}

\begin{wraptable}{r}{0.58\textwidth}
    \centering
    \small
    \setlength{\tabcolsep}{3pt}
    \caption{ImageNet-100 transfer from DINOv2 at IPC$=$1.}
    \label{tab:transfer_controlled}
    \begin{tabular}{lcc}
        \toprule
        Method & Source & Unseen \\
        \midrule
        Frozen-feature KIP & $84.5\pm0.3$ & 43.8 \\
        Frozen-feature FRePo & $86.3\pm0.2$ & 46.7 \\
        CLP-DD w/ MSE & $88.3\pm0.0$ & 37.5 \\
        \textbf{CLP-DD} & $\mathbf{89.2\pm0.1}$ & \textbf{50.9} \\
        \bottomrule
    \end{tabular}
\end{wraptable}

\textbf{Cross-backbone transfer.}
Under a matched protocol, we test DINOv2-distilled ImageNet-100 sets (IPC$=1$) on CLIP, MoCo-v3, and EVA-02. In Table~\ref{tab:transfer_controlled}, CLP-DD gains 7.1 and 4.2 points over frozen-feature KIP and FRePo. MSE lowers the unseen mean by 13.4 points. Accuracy drops from 89.2\% on DINOv2 to 50.9\% on unseen backbones, reflecting representation dependence.

\WFclear
\noindent\begin{minipage}[t]{0.45\textwidth}
    \vspace{0pt}
\textbf{ImageNet-1K efficiency.}
Figure~\ref{fig:imagenet1k_efficiency} shows that CLP-DD attains $75.6\pm0.1$\% accuracy at 46.33\,s per 100 iterations and 11,924\,MB peak memory. All methods use matched hardware, backbone, and batch settings. Timings cover complete distillation iterations and exclude setup and final evaluation. Relative to LGM with DSA ($m=3$), it provides a 13.9$\times$ speedup and 87.6\% lower peak memory with comparable accuracy. Against DSA-free LGM, it gains 12.4 accuracy points for 11.6\% more time and 17.7\% more memory. These results highlight the practical value of CLP-DD for large-scale frozen-backbone adaptation when both high linear-probe accuracy and limited computational resources are important. 
\end{minipage}\hfill
\begin{minipage}[t]{0.5\textwidth}
    \vspace{0pt}
    \centering
    \includegraphics[width=\linewidth]{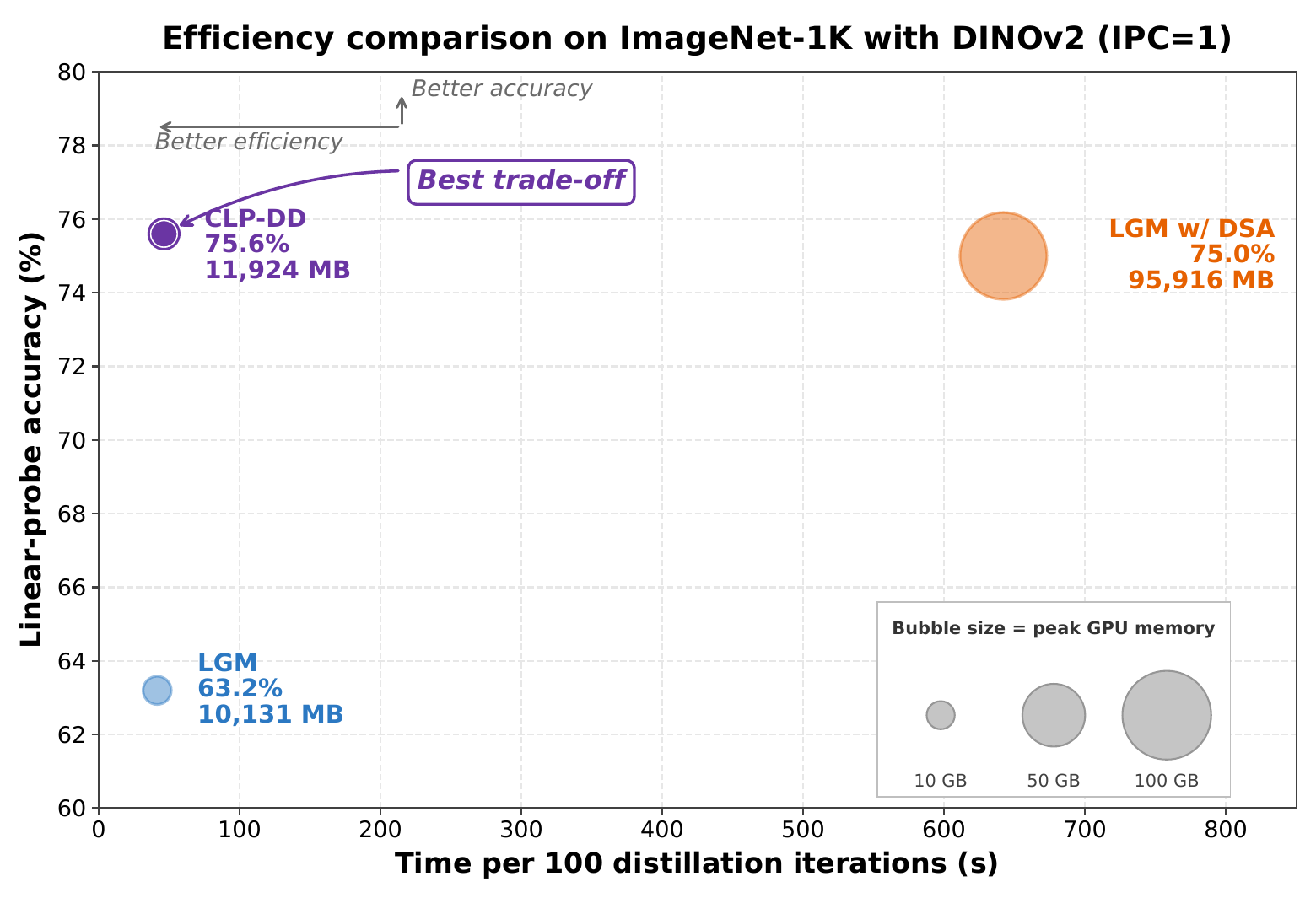}
    \captionof{figure}{ImageNet-1K/DINOv2, IPC$=$1, on a 96\,GB RTX PRO 6000. Bubble area: peak memory. Accuracy: three-run mean. Time excludes setup and evaluation.}
    \label{fig:imagenet1k_efficiency}
\end{minipage}\par

\WFclear
\endgroup
\section{Discussion}
\label{sec:discussion}

CLP-DD targets low-IPC adaptation to a specified frozen encoder. Centroid avoids iterative distillation and becomes competitive at higher IPC, trailing CLP-DD by 0.9 points on DINOv2 at IPC$=$10 (Appendix~\ref{app:ipc_scaling}). CLP-DD saves computation relative to LGM with DSA, while its advantage over DSA-free LGM is primarily accuracy at comparable or moderately higher cost.

The transfer results support the class-anchor objective beyond the source encoder, but the substantial matched-to-unseen-backbone drop limits portability. Same-backbone adaptation remains the setting with the strongest evidence; the full matrix and target-wise ablation are provided in Appendices~\ref{app:cross_backbone} and~\ref{subsec:additional_validation}. On Spawrious, CLP-DD is competitive with real-image selection (47.9\% versus 48.4\% for Centroid), showing that stronger in-distribution linear probing does not ensure robustness to background shift. Distillation may retain biases of the frozen representation, motivating complementary encoders and robustness-oriented objectives.

Broader probe losses, nonlinear heads, and partial fine-tuning remain untested. Positive ridge regularization guarantees invertibility, while acceptable finite-precision behavior is supported only over the tested range (Appendix~\ref{app:numerical_stability}). The mild late-stage decline remains unexplained. Final CLP-DD accuracies use a fixed 4,000-iteration budget without validation- or test-based checkpoint selection; training-only stopping and additional regularization remain future directions. With three runs, small accuracy differences warrant descriptive interpretation rather than significance claims.

\section{Conclusion}
\label{conclusion}

In this work, we introduced Closed-Form Linear-Probe Dataset Distillation (CLP-DD), a distillation framework tailored to frozen-backbone linear probing with pre-trained vision models. CLP-DD exploits the closed-form structure of ridge-regularized linear-probe adaptation in sample-space kernel form, enabling efficient low-IPC distillation without unrolled inner-loop optimization. Combined with a temperature-scaled class-anchor cross-entropy objective, it updates synthetic images to induce discriminative classifiers in the frozen feature space. Experiments on ImageNet-100 and ImageNet-1K with multiple pre-trained backbones show that CLP-DD improves average accuracy over real-image baselines at IPC$=$1 and serves as an efficient DSA-free alternative to LGM, while remaining competitive on spurious-correlation benchmarks. Limitations and broader impacts are discussed in Appendices~\ref{app:limitation} and~\ref{app:impact}.

\section*{Acknowledgments}

This research was supported in part by JSPS KAKENHI Grant Numbers JP24K02942 and JP25K21218.

\bibliographystyle{plain} 
\bibliography{ref}

@inproceedings{DM,
  title={Dataset condensation with distribution matching},
  author={Zhao, Bo and Bilen, Hakan},
  booktitle={IEEE/CVF Winter Conference on Applications of Computer Vision},
  pages={6514--6523},
  year={2023}
}

@inproceedings{n2,
  title={Learning to generate synthetic training data using gradient matching and implicit differentiation},
  author={Medvedev, Dmitry and D'yakonov, Alexander},
  booktitle={International Conference on Analysis of Images, Social Networks and Texts},
  pages={138--150},
  year={2021}
}

@inproceedings{point2,
  title={Datadam: Efficient dataset distillation with attention matching},
  author={Sajedi, Ahmad and Khaki, Samir and Amjadian, Ehsan and Liu, Lucy Z and Lawryshyn, Yuri A and Plataniotis, Konstantinos N},
  booktitle={IEEE/CVF International Conference on Computer Vision},
  pages={17097--17107},
  year={2023}
}

@inproceedings{dsa,
  title={Dataset condensation with differentiable siamese augmentation},
  author={Zhao, Bo and Bilen, Hakan},
  booktitle={International Conference on Machine Learning},
  pages={12674--12685},
  year={2021}
}

@article{random,
  title={Super-samples from kernel herding},
  author={Chen, Yutian and Welling, Max and Smola, Alex},
  journal={arXiv preprint arXiv:1203.3472},
  year={2012}
}

@article{LD,
  title={Flexible dataset distillation: Learn labels instead of images},
  author={Bohdal, Ondrej and Yang, Yongxin and Hospedales, Timothy},
  journal={arXiv preprint arXiv:2006.08572},
  year={2020}
}

@inproceedings{DCC,
  title={Dataset condensation with contrastive signals},
  author={Lee, Saehyung and Chun, Sanghyuk and Jung, Sangwon and Yun, Sangdoo and Yoon, Sungroh},
  booktitle={International Conference on Machine Learning},
  pages={12352--12364},
  year={2022}
}

@article{yu2023review,
  title={Dataset Distillation: A Comprehensive Review},
  author={Yu, Ruonan and Liu, Songhua and Wang, Xinchao},
  journal={IEEE Transactions on Pattern Analysis and Machine Intelligence},
  volume={46},
  number={1},
  pages={150--170},
  year={2024},
  doi={10.1109/TPAMI.2023.3323376},
  publisher={IEEE}
}

@inproceedings{cui2023scaling,
  title={Scaling Up Dataset Distillation to ImageNet-1K with Constant Memory},
  author={Cui, Justin and Wang, Ruochen and Si, Si and Hsieh, Cho-Jui},
  booktitle={International Conference on Machine Learning},
  pages={6565--6590},
  year={2023}
}

@inproceedings{guo2024datm,
  title={Towards Lossless Dataset Distillation via Difficulty-Aligned Trajectory Matching},
  author={Guo, Ziyao and Wang, Kai and Cazenavette, George and Li, Hui and Zhang, Kaipeng and You, Yang},
  booktitle={International Conference on Learning Representations},
  year={2024}
}

@inproceedings{d4m,
  title={{D$^{4}$M}: Dataset Distillation via Disentangled Diffusion Model},
  author={Su, Duo and Hou, Junjie and Gao, Weizhi and Tian, Yingjie and Tang, Bowen},
  booktitle={IEEE/CVF Conference on Computer Vision and Pattern Recognition},
  pages={5809--5818},
  year={2024}
}

@misc{li2022awesome,
  author={Li, Guang and Zhao, Bo and Wang, Tongzhou},
  title={Awesome Dataset Distillation},
  howpublished={\url{https://github.com/Guang000/Awesome-Dataset-Distillation}},
  year={2022}
}

@article{KIP,
  title={Dataset distillation with infinitely wide convolutional networks},
  author={Nguyen, Timothy and Novak, Roman and Xiao, Lechao and Lee, Jaehoon},
  journal={Advances in Neural Information Processing Systems},
  volume={34},
  pages={5186--5198},
  year={2021}
}

@inproceedings{li2024generative,
  title={Generative Dataset Distillation: Balancing Global Structure and Local Details},
  author={Li, Longzhen and Li, Guang and Togo, Ren and Maeda, Keisuke and Ogawa, Takahiro and Haseyama, Miki},
  booktitle={IEEE/CVF Conference on Computer Vision and Pattern Recognition Workshops},
  pages={7664--7671},
  year={2024}
}

@inproceedings{li2025generative,
  title={Generative Dataset Distillation Based on Self-knowledge Distillation},
  author={Li, Longzhen and Li, Guang and Togo, Ren and Maeda, Keisuke and Ogawa, Takahiro and Haseyama, Miki},
  booktitle={IEEE International Conference on Acoustics, Speech and Signal Processing (ICASSP)},
  pages={1--5},
  year={2025},
  doi={10.1109/ICASSP49660.2025.10888970}
}

@inproceedings{ye2025igds,
  title={Information-Guided Diffusion Sampling for Dataset Distillation},
  author={Ye, Linfeng and Hamidi, Shayan Mohajer and Li, Guang and Ogawa, Takahiro and Haseyama, Miki and Plataniotis, Konstantinos N.},
  booktitle={Advances in Neural Information Processing Systems Workshops},
  year={2025}
}

@inproceedings{cazenavette2025dataset,
  title={Dataset Distillation for Pre-Trained Self-Supervised Vision Models},
  author={Cazenavette, George and Torralba, Antonio and Sitzmann, Vincent},
  booktitle={Advances in Neural Information Processing Systems (NeurIPS)},
  volume={38},
  year={2025},
  doi={10.52202/085713-2019}
}

@inproceedings{radford2021learning,
  title={Learning transferable visual models from natural language supervision},
  author={Radford, Alec and Kim, Jong Wook and Hallacy, Chris and Ramesh, Aditya and Goh, Gabriel and Agarwal, Sandhini and Sastry, Girish and Askell, Amanda and Mishkin, Pamela and Clark, Jack and others},
  booktitle={International Conference on Machine Learning (ICML)},
  pages={8748--8763},
  year={2021},
  organization={PMLR}
}

@article{oquab2024dinov2,
  title={{DINOv2}: Learning Robust Visual Features without Supervision},
  author={Oquab, Maxime and Darcet, Timoth{\'e}e and Moutakanni, Th{\'e}o and Vo, Huy and Szafraniec, Marc and Khalidov, Vasil and Fernandez, Pierre and Haziza, Daniel and Massa, Francisco and El-Nouby, Alaaeldin and Assran, Mahmoud and Ballas, Nicolas and Galuba, Wojciech and Howes, Russell and Huang, Po-Yao and Li, Shang-Wen and Misra, Ishan and Rabbat, Michael and Sharma, Vasu and Synnaeve, Gabriel and Xu, Hu and J{\'e}gou, Herv{\'e} and Mairal, Julien and Labatut, Patrick and Joulin, Armand and Bojanowski, Piotr},
  journal={Transactions on Machine Learning Research},
  year={2024}
}

@inproceedings{chen2021empirical,
  title={An empirical study of training self-supervised vision transformers},
  author={Chen, Xinlei and Xie, Saining and He, Kaiming},
  booktitle={Proceedings of the IEEE/CVF International Conference on Computer Vision (ICCV)},
  pages={9640--9649},
  year={2021}
}

@article{fang2023eva,
  title={{EVA-02}: A Visual Representation for Neon Genesis},
  author={Fang, Yuxin and Sun, Quan and Wang, Xinggang and Huang, Tiejun and Wang, Xinlong and Cao, Yue},
  journal={arXiv preprint arXiv:2303.11331},
  year={2023}
}

@article{wang2018dataset,
  title={Dataset distillation},
  author={Wang, Tongzhou and Zhu, Jun-Yan and Torralba, Antonio and Efros, Alexei A},
  journal={arXiv preprint arXiv:1811.10959},
  year={2018}
}

@inproceedings{zhao2020dataset,
  title={Dataset condensation with gradient matching},
  author={Zhao, Bo and Mopuri, Konda Reddy and Bilen, Hakan},
  booktitle={International Conference on Learning Representations (ICLR)},
  year={2021}
}

@inproceedings{cazenavette2022dataset,
  title={Dataset distillation by matching training trajectories},
  author={Cazenavette, George and Wang, Tongzhou and Torralba, Antonio and Efros, Alexei A and Zhu, Jun-Yan},
  booktitle={Proceedings of the IEEE/CVF Conference on Computer Vision and Pattern Recognition (CVPR)},
  pages={4750--4759},
  year={2022}
}

@inproceedings{wang2022cafe,
  title={Cafe: Learning to condense dataset by aligning features},
  author={Wang, Kai and Zhao, Bo and Peng, Xiangyu and Zhu, Zheng and Yang, Shuo and Wang, Shuo and Huang, Guan and Bilen, Hakan and Wang, Xinchao and You, Yang},
  booktitle={Proceedings of the IEEE/CVF Conference on Computer Vision and Pattern Recognition (CVPR)},
  pages={12196--12205},
  year={2022}
}

@inproceedings{nguyen2020dataset,
  title={Dataset meta-learning from kernel ridge-regression},
  author={Nguyen, Timothy and Chen, Zhourong and Lee, Jaehoon},
  booktitle={International Conference on Learning Representations (ICLR)},
  year={2021}
}

@inproceedings{zhou2022dataset,
  title={Dataset distillation using neural feature regression},
  author={Zhou, Yongchao and Nezhadarya, Ehsan and Ba, Jimmy},
  booktitle={Advances in Neural Information Processing Systems (NeurIPS)},
  volume={35},
  pages={9813--9827},
  year={2022}
}

@inproceedings{rajeswaran2019meta,
  title={Meta-learning with implicit gradients},
  author={Rajeswaran, Aravind and Finn, Chelsea and Kakade, Sham M and Levine, Sergey},
  booktitle={Advances in Neural Information Processing Systems (NeurIPS)},
  volume={32},
  pages={113--124},
  year={2019}
}

@inproceedings{bai2019deep,
  title={Deep Equilibrium Models},
  author={Bai, Shaojie and Kolter, J. Zico and Koltun, Vladlen},
  booktitle={Advances in Neural Information Processing Systems},
  volume={32},
  pages={688--699},
  year={2019}
}

@article{russakovsky2015imagenet,
  title={Imagenet large scale visual recognition challenge},
  author={Russakovsky, Olga and Deng, Jia and Su, Hao and Krause, Jonathan and Satheesh, Sanjeev and Ma, Sean and Huang, Zhiheng and Karpathy, Andrej and Khosla, Aditya and Bernstein, Michael and others},
  journal={International Journal of Computer Vision (IJCV)},
  volume={115},
  number={3},
  pages={211--252},
  year={2015},
  publisher={Springer}
}

@inproceedings{tian2020contrastive,
  title={Contrastive multiview coding},
  author={Tian, Yonglong and Krishnan, Dilip and Isola, Phillip},
  booktitle={European Conference on Computer Vision (ECCV)},
  pages={776--794},
  year={2020},
  organization={Springer}
}

@inproceedings{sagawa2019distributionally,
  title={Distributionally robust neural networks for group shifts: On the importance of regularization for worst-case generalization},
  author={Sagawa, Shiori and Koh, Pang Wei and Hashimoto, Tatsunori B and Liang, Percy},
  booktitle={International Conference on Learning Representations (ICLR)},
  year={2020}
}

@article{hayes2023spawrious,
  title={Spawrious: A Benchmark for Fine Control of Spurious Correlation Biases},
  author={Lynch, Aengus and Dovonon, Gb{\`e}tondji J-S and Kaddour, Jean and Silva, Ricardo},
  journal={arXiv preprint arXiv:2303.05470},
  year={2023}
}

@inproceedings{ran2026tgdd,
  title={{TGDD}: Trajectory Guided Dataset Distillation with Balanced Distribution},
  author={Ran, Fengli and Pu, Xiao and Liu, Bo and Bi, Xiuli and Xiao, Bin},
  booktitle={Proceedings of the AAAI Conference on Artificial Intelligence (AAAI)},
  pages={15662--15670},
  year={2026},
  doi={10.1609/aaai.v40i18.38596}
}

@inproceedings{li2025hdd,
  title={Hyperbolic Dataset Distillation},
  author={Li, Wenyuan and Li, Guang and Maeda, Keisuke and Ogawa, Takahiro and Haseyama, Miki},
  booktitle={Proceedings of the Advances in Neural Information Processing Systems (NeurIPS)},
  year={2025}
}

@inproceedings{chansantiago2026learnability,
  title={Learnability-Guided Diffusion for Dataset Distillation},
  author={Chan-Santiago, Jeffrey A. and Shah, Mubarak},
  booktitle={Proceedings of the IEEE/CVF Conference on Computer Vision and Pattern Recognition (CVPR)},
  pages={41657--41666},
  year={2026}
}

@inproceedings{kim2022dataset,
  title={Dataset Condensation via Efficient Synthetic-Data Parameterization},
  author={Kim, Jang-Hyun and Kim, Jinuk and Oh, Seong Joon and Yun, Sangdoo and Song, Hwanjun and Jeong, Joonhyun and Ha, Jung-Woo and Song, Hyun Oh},
  booktitle={Proceedings of the International Conference on Machine Learning (ICML)},
  pages={11102--11118},
  year={2022}
}

@inproceedings{deng2022remember,
  title={Remember the Past: Distilling Datasets into Addressable Memories for Neural Networks},
  author={Deng, Zhiwei and Russakovsky, Olga},
  booktitle={Proceedings of the Advances in Neural Information Processing Systems (NeurIPS)},
  year={2022}
}

@article{jiang2025gsdd,
  title={Beyond Pixels: Efficient Dataset Distillation via Sparse Gaussian Representation},
  author={Jiang, Chenyang and Li, Zhengcen and Zhao, Hang and Shan, Qiben and Wu, Shaocong and Su, Jingyong},
  journal={arXiv preprint arXiv:2509.26219},
  year={2025}
}

@inproceedings{tran2026ptqdc,
  title={Post Training Quantization for Efficient Dataset Condensation},
  author={Tran, Linh-Tam and Bae, Sung-Ho},
  booktitle={Proceedings of the AAAI Conference on Artificial Intelligence (AAAI)},
  pages={25992--26000},
  year={2026},
  doi={10.1609/aaai.v40i31.39800}
}

@inproceedings{liu2025wasserstein,
  title={Dataset Distillation via the Wasserstein Metric}, 
  author={Liu, Haoyang and Li, Yijiang and Xing, Tiancheng and Wang, Peiran and Dalal, Vibhu and Li, Luwei and He, Jingrui and Wang, Haohan},
  booktitle={Proceedings of the IEEE/CVF International Conference on Computer Vision (ICCV)},
  year={2025}
}

@inproceedings{lee2024selmatch,
  title={{SelMatch}: Effectively Scaling Up Dataset Distillation via Selection-Based Initialization and Partial Updates by Trajectory Matching}, 
  author={Lee, Yongmin and Chung, Hye Won},
  booktitle={Proceedings of the International Conference on Machine Learning (ICML)},
  pages={26546--26567},
  year={2024}
}

@inproceedings{zhong2025mct,
  title={Towards Stable and Storage-efficient Dataset Distillation: Matching Convexified Trajectory},
  author={Zhong, Wenliang and Tang, Haoyu and Zheng, Qinghai and Xu, Mingzhu and Hu, Yupeng and Guan, Weili},
  booktitle={Proceedings of the IEEE/CVF Conference on Computer Vision and Pattern Recognition (CVPR)},
  pages={25581--25589},
  year={2025}
}

@inproceedings{loo2022efficient,
  title={Efficient Dataset Distillation using Random Feature Approximation},
  author={Loo, Noel and Hasani, Ramin and Amini, Alexander and Rus, Daniela},
  booktitle={Proceedings of the Advances in Neural Information Processing Systems (NeurIPS)},
  year={2022}
}

\newpage
\raggedbottom
\appendix

\section{Detailed Mathematical Derivations}
\label{app:detailed_derivations}

This appendix provides the derivations used in Closed-Form Linear-Probe Dataset Distillation (CLP-DD). We first derive the closed-form ridge solution and its steady-state interpretation under gradient descent. We then present the sample-space kernel reformulation, the differential through the closed-form solver, the gradient of the outer class-anchor objective, and the final gradient flow to the synthetic images.

\subsection{Notation and Ridge Inner Problem}
\label{app:notation}

For compactness in the appendix, we drop the `syn' subscript
\begin{equation}
X := X_{\mathrm{syn}} \in \mathbb{R}^{N\times d},
\qquad
Y := Y_{\mathrm{syn}} \in \mathbb{R}^{N\times C},
\end{equation}
where $N$ is the number of synthetic samples, $d$ is the feature dimension, and $C$ is the number of classes. The linear probe is denoted by $W\in\mathbb{R}^{d\times C}$.

The ridge-regularized inner objective is
\begin{equation}
\mathcal{L}_{\mathrm{in}}(W)
=
\frac{1}{2}\|XW-Y\|_F^2
+
\frac{\lambda}{2}\|W\|_F^2,
\qquad
\lambda>0.
\label{eq:app_inner_loss}
\end{equation}
Using standard matrix calculus, its gradient is
\begin{equation}
\nabla_W \mathcal{L}_{\mathrm{in}}(W)
=
(X^\top X+\lambda I_d)W-X^\top Y.
\label{eq:app_inner_grad}
\end{equation}
Defining
\begin{equation}
H=X^\top X+\lambda I_d,
\qquad
c=X^\top Y,
\label{eq:app_H_c}
\end{equation}
we can write Eq.~\eqref{eq:app_inner_grad} as
\begin{equation}
\nabla_W \mathcal{L}_{\mathrm{in}}(W)=HW-c.
\end{equation}
Since $\lambda>0$, $H$ is symmetric positive definite and hence invertible. Setting the gradient to zero gives the standard ridge-regression solution
\begin{equation}
W^*
=
H^{-1}c
=
(X^\top X+\lambda I_d)^{-1}X^\top Y.
\label{eq:app_primal_solution}
\end{equation}

\subsection{Steady-State View of Ridge-Regression Linear Probing}
\label{app:steady_state}

The closed-form solution in Eq.~\eqref{eq:app_primal_solution} can also be viewed as the steady state of gradient descent on the ridge-regularized linear-probe objective. Gradient descent with step size $\eta>0$ follows
\begin{equation}
W_{t+1}
=
W_t-\eta(HW_t-c)
=
(I_d-\eta H)W_t+\eta c.
\label{eq:app_gd_update}
\end{equation}
Let
\begin{equation}
B=I_d-\eta H.
\end{equation}
Starting from $W_0=\mathbf{0}$, unrolling the recursion gives
\begin{equation}
W_t
=
\eta\sum_{k=0}^{t-1}B^k c.
\end{equation}
Using the finite matrix geometric series and $I_d-B=\eta H$, we obtain
\begin{equation}
W_t
=
\eta(I_d-B^t)(I_d-B)^{-1}c
=
(I_d-B^t)H^{-1}c.
\label{eq:app_Wt}
\end{equation}
If the step size satisfies the standard stability condition
\begin{equation}
0<\eta<\frac{2}{\mu_{\max}(H)},
\label{eq:app_stability}
\end{equation}
then $\rho(B)<1$ and $B^t\to 0$. Therefore,
\begin{equation}
\lim_{t\to\infty}W_t
=
H^{-1}c
=
(X^\top X+\lambda I_d)^{-1}X^\top Y.
\end{equation}
Thus, the unrolled linear-probe dynamics converge to the same classifier as the closed-form ridge solution used by CLP-DD.

\subsection{Sample-Space Kernel Reformulation}
\label{app:kernel_reformulation}

We next derive the sample-space form of the ridge solution. For $\lambda>0$, both $X^\top X+\lambda I_d$ and $XX^\top+\lambda I_N$ are invertible. We have
\begin{equation}
(X^\top X+\lambda I_d)X^\top
=
X^\top(XX^\top+\lambda I_N).
\end{equation}
Multiplying the two sides on the left by $(X^\top X+\lambda I_d)^{-1}$ and on the right by $(XX^\top+\lambda I_N)^{-1}$ gives the standard push-through identity
\begin{equation}
(X^\top X+\lambda I_d)^{-1}X^\top
=
X^\top(XX^\top+\lambda I_N)^{-1}.
\label{eq:app_push_through}
\end{equation}
Substituting Eq.~\eqref{eq:app_push_through} into Eq.~\eqref{eq:app_primal_solution} yields
\begin{equation}
W^*
=
X^\top(XX^\top+\lambda I_N)^{-1}Y.
\label{eq:app_kernel_solution}
\end{equation}
Let
\begin{equation}
K=XX^\top,
\qquad
A=K+\lambda I_N.
\end{equation}
Then the sample-space solver used in CLP-DD is
\begin{equation}
W^*
=
X^\top A^{-1}Y.
\label{eq:app_kernel_solver}
\end{equation}
This replaces a $d\times d$ linear system with an $N\times N$ system, which is suitable for the low-IPC regime where the number of synthetic samples is small. When $N$ is not smaller than $d$, the primal form in Eq.~\eqref{eq:app_primal_solution} can be used instead without changing the induced classifier.

\subsection{Differentiating Through the Closed-Form Solver}
\label{app:kernel_gradient}

We now derive the differential through the sample-space solver. This derivation clarifies how the outer objective depends on the synthetic features.

Let
\begin{equation}
A=XX^\top+\lambda I_N,
\qquad
S=A^{-1},
\qquad
P=SY=A^{-1}Y.
\end{equation}
The kernel-space solution is
\begin{equation}
W^*=X^\top P.
\end{equation}
For fixed $Y$, its differential with respect to $X$ is
\begin{equation}
\mathrm{d}W^*
=
\mathrm{d}X^\top P
-
X^\top S\mathrm{d}X X^\top P
-
X^\top SX\mathrm{d}X^\top P.
\label{eq:app_dW_kernel}
\end{equation}
This follows by differentiating $W^*=X^\top P$:
\begin{equation}
\mathrm{d}W^*
=
\mathrm{d}X^\top P+X^\top \mathrm{d}P.
\end{equation}
Since $P=A^{-1}Y$ and $Y$ is fixed,
\begin{equation}
\mathrm{d}P
=
\mathrm{d}(A^{-1})Y
=
-A^{-1}(\mathrm{d}A)A^{-1}Y
=
-S(\mathrm{d}A)P,
\end{equation}
where
\begin{equation}
\mathrm{d}(A^{-1})
=
-A^{-1}(\mathrm{d}A)A^{-1}.
\end{equation}
Moreover,
\begin{equation}
\mathrm{d}A
=
\mathrm{d}X X^\top + X\mathrm{d}X^\top.
\end{equation}
Substituting these expressions into $\mathrm{d}W^*$ gives Eq.~\eqref{eq:app_dW_kernel}.

Let
\begin{equation}
G=
\frac{\partial \mathcal{L}_{\mathrm{meta}}}{\partial W^*}
\in\mathbb{R}^{d\times C}
\end{equation}
be the upstream gradient from the outer objective. Then
\begin{equation}
\mathrm{d}\mathcal{L}_{\mathrm{meta}}
=
\operatorname{tr}(G^\top\mathrm{d}W^*).
\end{equation}
Using Eq.~\eqref{eq:app_dW_kernel} and collecting terms in the form
$\operatorname{tr}((\nabla_X\mathcal{L}_{\mathrm{meta}})^\top\mathrm{d}X)$ gives
\begin{equation}
\nabla_X \mathcal{L}_{\mathrm{meta}}
=
P G^\top
-
S XG P^\top X
-
P G^\top X^\top S X.
\label{eq:app_grad_X_kernel}
\end{equation}
In practice, CLP-DD computes this gradient using automatic differentiation through a linear solve, but Eq.~\eqref{eq:app_grad_X_kernel} makes explicit how the synthetic features influence the outer objective through the closed-form classifier.

\subsection{Outer Objective Gradient}
\label{app:outer_loss_gradient}

The outer objective is a temperature-scaled cross-entropy loss over the closed-form classifier:
\begin{equation}
\mathcal{L}_{\mathrm{meta}}
=
-\frac{1}{M}
\sum_{i=1}^{M}
\log
\frac{
\exp(x_{\mathrm{real},i}^\top w^*_{y_i}/\tau)
}{
\sum_{j=1}^{C}
\exp(x_{\mathrm{real},i}^\top w^*_{j}/\tau)
}.
\label{eq:app_meta_loss}
\end{equation}
Let
\begin{equation}
X_{\mathrm{real}}\in\mathbb{R}^{M\times d},
\qquad
T\in\mathbb{R}^{M\times C}
\end{equation}
denote the real feature matrix and the corresponding one-hot label matrix. Define
\begin{equation}
Z=\frac{1}{\tau}X_{\mathrm{real}}W^*,
\qquad
\Pi=\operatorname{softmax}(Z),
\end{equation}
where the softmax is applied row-wise over the $C$ classes. The loss can be written as
\begin{equation}
\mathcal{L}_{\mathrm{meta}}
=
-\frac{1}{M}
\sum_{i=1}^{M}
\sum_{j=1}^{C}
T_{ij}\log\Pi_{ij}.
\end{equation}
Using the standard softmax-cross-entropy derivative,
\begin{equation}
\frac{\partial \mathcal{L}_{\mathrm{meta}}}{\partial Z}
=
\frac{1}{M}(\Pi-T).
\end{equation}
Since $Z=\tau^{-1}X_{\mathrm{real}}W^*$, the gradient with respect to the closed-form classifier is
\begin{equation}
G
=
\frac{\partial \mathcal{L}_{\mathrm{meta}}}{\partial W^*}
=
\frac{1}{M\tau}
X_{\mathrm{real}}^\top(\Pi-T).
\label{eq:app_grad_W_outer}
\end{equation}
Substituting Eq.~\eqref{eq:app_grad_W_outer} into Eq.~\eqref{eq:app_grad_X_kernel} gives the gradient of the outer objective with respect to the synthetic features.

\subsection{Gradient Flow to Synthetic Images}
\label{app:synthetic_image_gradient}

The synthetic features are produced by the frozen backbone:
\begin{equation}
X_{\mathrm{syn}}
=
\phi(\mathcal{S}_{\mathrm{syn}}).
\end{equation}
Although the parameters of $\phi$ are fixed, the synthetic images remain learnable. For the $n$-th synthetic image $s_n$, with feature $x_n=\phi(s_n)$, the gradient is
\begin{equation}
\nabla_{s_n}\mathcal{L}_{\mathrm{meta}}
=
J_\phi(s_n)^\top
\nabla_{x_n}\mathcal{L}_{\mathrm{meta}},
\end{equation}
where
\begin{equation}
J_\phi(s_n)
=
\frac{\partial \phi(s_n)}{\partial s_n}
\end{equation}
is the Jacobian of the frozen feature extractor with respect to its input. Stacking all the synthetic samples gives
\begin{equation}
\nabla_{\mathcal{S}_{\mathrm{syn}}}\mathcal{L}_{\mathrm{meta}}
=
\left(
\frac{\partial X_{\mathrm{syn}}}{\partial \mathcal{S}_{\mathrm{syn}}}
\right)^\top
\nabla_{X_{\mathrm{syn}}}\mathcal{L}_{\mathrm{meta}}.
\label{eq:app_grad_images}
\end{equation}
Thus, CLP-DD avoids differentiating through an unrolled inner trajectory and instead propagates gradients through the closed-form linear-probe solution and the frozen feature extractor.

\section{Implementation Details}
\label{app:implementation_details}

This section provides additional implementation details on datasets, backbone models, distillation settings, evaluation protocols, and data augmentations.

\subsection{Datasets}

We use the same four datasets and splits as LGM~\cite{cazenavette2025dataset}.

\paragraph{ImageNet-1K.} We use the ILSVRC 2012 classification release~\cite{russakovsky2015imagenet}, with the official training split for distillation and the validation split for evaluation. The \href{https://www.image-net.org/challenges/LSVRC/2012/2012-downloads.php}{ILSVRC image terms} limit use to non-commercial research and education.

\paragraph{ImageNet-100.} We restrict both ILSVRC 2012 splits to the 100 WordNet synsets from CMC~\cite{tian2020contrastive}, using the \href{https://github.com/GeorgeCazenavette/linear-gradient-matching/blob/main/src/data/dataloaders/imagenet100.txt}{class list in the LGM implementation}. The subset is subject to the same ILSVRC image terms.

\paragraph{Spawrious.} We use the official One-to-One Hard split (\texttt{o2o\_hard})~\cite{hayes2023spawrious}, distributed as \texttt{spawrious224\_\_o2o\_hard.tar.gz}. The training split pairs four dog breeds with different backgrounds, and the official test split changes those pairings. The upstream \href{https://github.com/aengusl/spawrious\#licensing}{README lists CC BY 4.0}. The repository \href{https://github.com/aengusl/spawrious/blob/main/LICENSE}{LICENSE file is CC0-1.0}.

\paragraph{WaterBirds.} We use Waterbirds v1.0~\cite{sagawa2019distributionally} through the \href{https://github.com/p-lambda/wilds/blob/main/wilds/datasets/waterbirds_dataset.py}{WILDS loader}, with the official training and test splits defined in \texttt{metadata.csv}. The images combine CUB-200-2011 birds with Places backgrounds. The dataset documentation limits use to non-commercial research and education. The \href{https://github.com/p-lambda/wilds/blob/main/LICENSE}{WILDS software license} is MIT.

\subsection{Backbone Models}

We use the four ViT-B checkpoints from the \href{https://github.com/GeorgeCazenavette/linear-gradient-matching/blob/main/src/models/__init__.py}{LGM implementation}~\cite{cazenavette2025dataset}. We extract final-layer features and freeze the backbone parameters throughout distillation and evaluation.

\paragraph{CLIP.} We use OpenAI \texttt{ViT-B/16}~\cite{radford2021learning} from the \href{https://github.com/openai/CLIP/blob/main/LICENSE}{MIT-licensed CLIP repository}.

\paragraph{DINOv2.} We use Torch Hub \texttt{dinov2\_vitb14}~\cite{oquab2024dinov2}. The \href{https://github.com/facebookresearch/dinov2\#license}{official release} licenses the code and weights under Apache-2.0.

\paragraph{EVA-02.} We use \texttt{eva02\_base\_patch14\_224.mim\_in22k}~\cite{fang2023eva}. The \href{https://huggingface.co/timm/eva02_base_patch14_224.mim_in22k}{checkpoint card lists MIT}, and the \href{https://github.com/huggingface/pytorch-image-models/blob/main/LICENSE}{\texttt{timm} library} uses Apache-2.0.

\paragraph{MoCo-v3.} We use \href{https://huggingface.co/nyu-visionx/moco-v3-vit-b}{\texttt{nyu-visionx/moco-v3-vit-b}}~\cite{chen2021empirical}. The \href{https://github.com/facebookresearch/moco-v3/blob/main/LICENSE}{upstream Meta repository} uses CC BY-NC 4.0. The Hugging Face distribution does not specify a separate license.

\subsection{Distillation Settings}

Synthetic images are randomly initialized and optimized for $4,000$ distillation iterations. Unless otherwise specified, the image resolution is fixed to $224\times224$, the ridge coefficient is set to $\lambda=0.1$, and the temperature in the outer objective is set to $\tau=0.07$. At each distillation iteration, we sample a class-balanced real batch with $B=4$ real samples per class to compute the outer loss.

The synthetic images are optimized using Adam with an initial learning rate of $0.05$ and a cosine annealing schedule. Gradients are backpropagated through the frozen feature extractor and the closed-form linear-probe solver, while the backbone parameters remain fixed. Unless otherwise specified, the batch size for linear-probe training is set to $256$.

The experiments use NVIDIA RTX 5090 GPUs and NVIDIA RTX Pro 6000 GPUs with 96GB memory. ImageNet-100 efficiency profiling is conducted on an RTX 5090, while the additional higher-IPC accuracy experiments and the ImageNet-1K efficiency profile use an RTX Pro 6000. Within each timing comparison, all methods use the same GPU model, backbone, image resolution, and batch configuration; runtime is not compared across GPU models.

\subsection{Evaluation Protocol}

After distillation, we evaluate the synthetic set by training a randomly initialized linear classifier on top of frozen backbone features. The linear classifier is trained only on the distilled images and evaluated on the ImageNet validation splits or on the official test splits for Spawrious and WaterBirds. Unless otherwise specified, the classifier is trained for $500$ epochs using Adam with a learning rate of $0.01$.

For real-image baselines, we use the same frozen backbone and the same linear-probe evaluation protocol. This ensures that the comparison reflects the quality of the distilled or selected images, rather than differences in classifier training or feature extraction.

\subsection{Augmentation}

By default, CLP-DD uses standard image augmentations, including random cropping and noise injection, while keeping the spatial resolution fixed at $224\times224$. CLP-DD does not use Differentiable Siamese Augmentation (DSA) by default. DSA is used only for explicitly identified LGM configurations and the optional CLP-DD augmentation ablation.

\section{Additional Experimental Results}
\label{app:additional_experiments}

This section provides the additional cross-backbone, IPC-scaling, efficiency, augmentation, hyperparameter-sensitivity, and numerical-stability results. Accuracy results are reported as mean $\pm$ standard deviation over three independent runs unless indicated otherwise.

\subsection{Cross-Backbone Transfer}
\label{app:cross_backbone}

Table~\ref{tab:app_transfer} evaluates each ImageNet-100 synthetic set at IPC$=$1 with all four frozen backbones. Diagonal entries use the same backbone for distillation and evaluation; the unseen average excludes the diagonal. The unseen averages of 50.9--52.3\% are substantially below matched-backbone accuracy, demonstrating limited transfer and substantial source-representation dependence rather than backbone invariance.

\begin{table}[htbp]
\centering
\small
\setlength{\tabcolsep}{6pt}
\caption{Cross-backbone transfer on ImageNet-100 at IPC$=$1. Rows identify the distillation backbone and columns the evaluation backbone. Individual accuracies are mean $\pm$ standard deviation over three runs; unseen averages report the mean over the three off-diagonal backbones.}
\label{tab:app_transfer}
\begin{tabular}{lccccc}
\toprule
Source & DINOv2 & MoCo-v3 & CLIP & EVA-02 & Unseen avg. \\
\midrule
DINOv2 & $\mathbf{89.2\pm0.1}$ & $52.0\pm0.5$ & $46.8\pm0.6$ & $54.0\pm0.4$ & 50.9 \\
MoCo-v3 & $57.2\pm0.2$ & $\mathbf{83.9\pm0.1}$ & $45.6\pm0.3$ & $54.1\pm0.1$ & 52.3 \\
CLIP & $53.8\pm0.1$ & $46.5\pm0.1$ & $\mathbf{79.2\pm0.1}$ & $55.3\pm0.3$ & 51.9 \\
EVA-02 & $57.0\pm0.1$ & $48.4\pm0.1$ & $51.2\pm0.2$ & $\mathbf{87.6\pm0.0}$ & 52.2 \\
\bottomrule
\end{tabular}
\end{table}

\subsection{Controlled Cross-Backbone Comparisons}
\label{subsec:additional_validation}

With DINOv2 as the source, Table~\ref{tab:transfer_controlled} compares methods under the same fixed features, resolution, IPC, optimization budget, and evaluation protocol. CLP-DD improves the unseen average by 7.1 points over frozen-feature KIP and 4.2 points over frozen-feature FRePo. FRePo's evolving model pool is inapplicable to the fixed-encoder setting. Replacing the class-anchor objective with MSE reduces unseen accuracy by 13.4 points, providing additional evidence for the classification-oriented outer loss. Same-backbone frozen-feature adaptation remains the primary use case.

Table~\ref{tab:transfer_objective} gives the individual target-backbone results and the reported three-run unseen averages for the objective ablation. This changes the distillation-stage objective, rather than the downstream evaluation objective, and therefore does not establish robustness to substantially different probe losses or nonlinear heads.

the \begin{table}[H]
\centering
\small
\setlength{\tabcolsep}{5pt}
\caption{Outer-objective ablation for transfer from DINOv2 on ImageNet-100 at IPC$=$1. Accuracy (\%) is mean $\pm$ standard deviation over three runs; the unseen average excludes DINOv2.}
\label{tab:transfer_objective}
\begin{tabular}{lccccc}
\toprule
Outer objective & DINOv2 & MoCo-v3 & CLIP & EVA-02 & Unseen avg. \\
\midrule
MSE & $88.3\pm0.0$ & $31.3\pm0.6$ & $39.6\pm0.4$ & $41.6\pm0.3$ & $37.5\pm0.4$ \\
Class-anchor & $\mathbf{89.2\pm0.1}$ & $\mathbf{52.0\pm0.5}$ & $\mathbf{46.8\pm0.6}$ & $\mathbf{54.0\pm0.4}$ & $\mathbf{50.9\pm0.5}$ \\
\bottomrule
\end{tabular}
\end{table}

\subsection{Extended IPC Scaling}
\label{app:ipc_scaling}

Table~\ref{tab:app_dino_scaling} extends the real-image-selection comparison to IPC$=\{1,2,3,5,10\}$ using DINOv2. CLP-DD remains more accurate than Neighbor, Centroid, Random, and DSA-free LGM at every tested budget. The margin over Centroid decreases from 2.3 to 0.9 points as IPC grows from 1 to 10. Thus, the largest advantage is at low IPC; the 0.9-point gain at IPC$=$10 requires 4,000 distillation iterations, whereas Centroid requires no iterative distillation.

\begin{table}[htbp]
\centering
\small
\setlength{\tabcolsep}{6pt}
\caption{Extended IPC scaling on ImageNet-100 with DINOv2. Accuracy (\%) is the mean $\pm$ standard deviation over three runs. LGM denotes the DSA-free variant.}
\label{tab:app_dino_scaling}
\begin{tabular}{lccccc}
\toprule
Method & IPC$=$1 & IPC$=$2 & IPC$=$3 & IPC$=$5 & IPC$=$10 \\
\midrule
Neighbor & $86.0\pm0.2$ & $87.8\pm0.3$ & $88.7\pm0.2$ & $89.7\pm0.2$ & $90.9\pm0.2$ \\
Centroid & $86.9\pm0.3$ & $88.5\pm0.2$ & $89.3\pm0.2$ & $90.2\pm0.2$ & $91.2\pm0.1$ \\
Random & $74.8\pm2.6$ & $79.0\pm2.4$ & $81.8\pm1.4$ & $84.2\pm1.3$ & $87.4\pm1.9$ \\
LGM & $82.6\pm0.2$ & $85.0\pm0.3$ & $86.3\pm0.2$ & $87.8\pm0.3$ & $89.6\pm0.4$ \\
\textbf{CLP-DD} & $\mathbf{89.2\pm0.1}$ & $\mathbf{90.7\pm0.1}$ & $\mathbf{91.3\pm0.1}$ & $\mathbf{91.8\pm0.0}$ & $\mathbf{92.1\pm0.1}$ \\
\bottomrule
\end{tabular}
\end{table}

Table~\ref{tab:app_moco_scaling} repeats the IPC$=\{1,5,10\}$ comparison with MoCo-v3. The margin over Centroid also narrows, from 6.2 to 3.8 points, but remains larger than on DINOv2. The diminishing-margin trend is shared by the two tested backbones, while the size of the improvement depends on the representation.

\begin{table}[htbp]
\centering
\small
\setlength{\tabcolsep}{10pt}
\caption{IPC scaling on ImageNet-100 with MoCo-v3. Accuracy (\%) is the mean $\pm$ standard deviation over three runs. LGM denotes the DSA-free variant.}
\label{tab:app_moco_scaling}
\begin{tabular}{lccc}
\toprule
Method & IPC$=$1 & IPC$=$5 & IPC$=$10 \\
\midrule
Centroid & $77.7\pm0.1$ & $81.7\pm0.2$ & $82.9\pm0.2$ \\
LGM & $59.5\pm0.5$ & $68.3\pm0.5$ & $73.3\pm0.4$ \\
\textbf{CLP-DD} & $\mathbf{83.9\pm0.1}$ & $\mathbf{86.4\pm0.1}$ & $\mathbf{86.7\pm0.1}$ \\
\bottomrule
\end{tabular}
\end{table}

\subsection{ImageNet-1K Efficiency}
\label{app:imagenet1k_efficiency}

Table~\ref{tab:imagenet1k_efficiency} extends profiling to ImageNet-1K with DINOv2 at IPC$=$1 on a 96 GB NVIDIA RTX PRO 6000. All methods use identical hardware and batch configuration within this comparison. LGM with DSA uses $m=3$ augmentation replicas. Time measures 100 complete distillation iterations, including method-specific feature extraction, augmentation, loss computation, backward propagation, synthetic-image updates, and the differentiable solve where applicable. Peak memory covers the same computation. One-time setup and final linear-probe evaluation are excluded for all methods.

\begin{table}[H]
\centering
\small
\setlength{\tabcolsep}{10pt}
\caption{ImageNet-1K/DINOv2 efficiency at IPC$=$1 on a 96\,GB RTX PRO 6000. Time is seconds per 100 complete distillation iterations, peak memory is in MB, and accuracy (\%) is the mean $\pm$ standard deviation over three runs.}
\label{tab:imagenet1k_efficiency}
\begin{tabular}{lrrc}
\toprule
Method & Time / 100 iters & Peak memory & Accuracy \\
\midrule
LGM w/ DSA ($m=3$) & 642.14 & 95,916 & $75.0\pm0.1$ \\
LGM & 41.51 & 10,131 & $63.2\pm0.6$ \\
\textbf{CLP-DD} & 46.33 & 11,924 & $\mathbf{75.6\pm0.1}$ \\
\bottomrule
\end{tabular}
\end{table}

CLP-DD is 13.9$\times$ faster than LGM with DSA and uses 12.4\% of its peak memory, while accuracy is comparable. Against DSA-free LGM, CLP-DD gains 12.4 accuracy points at 11.6\% more runtime and 17.7\% more peak memory. These results establish the accuracy--efficiency trade-off for this configuration; higher-IPC ImageNet-1K scaling remains untested. Absolute times are not compared with the ImageNet-100 RTX 5090 profile because the GPU models differ.

Extrapolating the measured optimization cost to 4,000 iterations gives about 1,853\,s for CLP-DD, 1,660\,s for DSA-free LGM, and 25,686\,s for LGM with DSA. These are estimates of the iterative optimization stage, excluding setup and evaluation, rather than measured end-to-end runtimes.

\subsection{Augmentation Ablations}
\label{app:augmentation_ablation}

Table~\ref{tab:augmentation_ablation} evaluates standard image augmentation and optional DSA on ImageNet-100 with DINOv2 at IPC$=$1. Removing cropping and noise reduces accuracy from $89.2\pm0.1$\% to $87.9\pm0.8$\%, still 5.3 points above DSA-free LGM. Adding DSA with multiplicity $m=1$ to CLP-DD yields 89.3\% accuracy, with no standard deviation reported for that configuration.

\begin{table}[H]
\centering
\small
\setlength{\tabcolsep}{8pt}
\caption{Augmentation ablations on ImageNet-100/DINOv2 at IPC$=$1. Accuracy (\%) is the mean $\pm$ standard deviation over three runs where available; the DSA accuracy has no reported standard deviation. Time is seconds per 100 complete iterations, and peak memory is in MB. A dash denotes an unreported profiling measurement.}
\label{tab:augmentation_ablation}
\begin{tabular}{lcrr}
\toprule
CLP-DD setting & Accuracy & Time / 100 iters & Peak memory \\
\midrule
Without image augmentation & $87.9\pm0.8$ & -- & -- \\
Default: cropping and noise & $89.2\pm0.1$ & 4.07 & 2,361 \\
Default w/ DSA ($m=1$) & $89.3\pm0.1$ & 9.39 & 4,011 \\
\bottomrule
\end{tabular}
\end{table}

Standard augmentation contributes 1.3 points, but does not alone explain CLP-DD's advantage over DSA-free LGM. Adding DSA changes accuracy by only 0.1 points while increasing runtime by 2.31$\times$ and peak memory by 1,650 MB. Under this tested configuration, CLP-DD therefore achieves strong performance without requiring DSA. Together with the MSE-objective ablation, these results support the role of the proposed objective, but provide only a partial decomposition: they do not uniquely attribute the remaining gap to LGM with DSA to any single component.

\subsection{Sensitivity Across Backbones}
\label{app:hyperparameter_sensitivity}

We vary one hyperparameter at a time on ImageNet-100 at IPC$=$1, fixing the other at its default. Tables~\ref{tab:app_temperature} and~\ref{tab:app_ridge} report the complete four-backbone sweeps, including the additional intermediate ridge coefficient $\lambda=0.20$. The same defaults, $\tau=0.07$ and $\lambda=0.10$, perform best among all tested values for every backbone. Accuracy decreases already at $\lambda=0.20$, with reductions of 2.1, 2.6, 1.6, and 1.7 points on DINOv2, MoCo-v3, CLIP, and EVA-02, respectively. No backbone-specific choice of the defaults is used.

\begin{table}[H]
\centering
\small
\setlength{\tabcolsep}{9pt}
\caption{Temperature sensitivity on ImageNet-100 at IPC$=$1 with $\lambda=0.10$. Accuracy (\%) is mean $\pm$ standard deviation from the three-run sensitivity sweep.}
\label{tab:app_temperature}
\begin{tabular}{lcccc}
\toprule
Backbone & $\tau=0.01$ & $\tau=0.05$ & $\tau=0.07$ & $\tau=0.10$ \\
\midrule
DINOv2 & $80.8\pm0.3$ & $86.7\pm0.2$ & $\mathbf{89.2\pm0.1}$ & $88.8\pm0.1$ \\
MoCo-v3 & $69.7\pm0.3$ & $78.1\pm0.2$ & $\mathbf{83.9\pm0.1}$ & $83.2\pm0.0$ \\
CLIP & $62.5\pm0.4$ & $73.9\pm0.3$ & $\mathbf{79.2\pm0.2}$ & $78.5\pm0.2$ \\
EVA-02 & $75.9\pm0.3$ & $83.9\pm0.2$ & $\mathbf{87.6\pm0.1}$ & $87.0\pm0.1$ \\
\bottomrule
\end{tabular}
\end{table}

\begin{table}[H]
\centering
\small
\setlength{\tabcolsep}{6pt}
\caption{Ridge-coefficient sensitivity on ImageNet-100 at IPC$=$1 with $\tau=0.07$. Accuracy (\%) is the mean $\pm$ standard deviation from the three-run sensitivity sweep; the $\lambda=0.20$ column reports the added intermediate setting.}
\label{tab:app_ridge}
\begin{tabular}{lccccc}
\toprule
Backbone & $\lambda=0.05$ & $\lambda=0.07$ & $\lambda=0.10$ & $\lambda=0.20$ & $\lambda=0.50$ \\
\midrule
DINOv2 & $86.8\pm0.2$ & $88.1\pm0.1$ & $\mathbf{89.2\pm0.1}$ & $87.1\pm0.2$ & $84.6\pm0.2$ \\
MoCo-v3 & $79.3\pm0.2$ & $81.3\pm0.2$ & $\mathbf{83.9\pm0.1}$ & $81.3\pm0.2$ & $76.2\pm0.1$ \\
CLIP & $74.8\pm0.3$ & $77.1\pm0.2$ & $\mathbf{79.2\pm0.2}$ & $77.6\pm0.3$ & $70.9\pm0.4$ \\
EVA-02 & $83.8\pm0.2$ & $85.9\pm0.2$ & $\mathbf{87.6\pm0.1}$ & $85.9\pm0.2$ & $80.9\pm0.3$ \\
\bottomrule
\end{tabular}
\end{table}

\subsection{Numerical Stability}
\label{app:numerical_stability}

We examine four backbones, three seeds, and IPC$=\{1,2,5,10\}$, for a total of 48 runs. For the regularized Gram matrix $A=X_{\mathrm{syn}}X_{\mathrm{syn}}^\top+\lambda I$, the median condition number increases from $7.4\times10^1$ at IPC$=$1 to $7.6\times10^2$ at IPC$=$10; the largest observed condition number is $6.4\times10^3$. The maximum relative solve residual remains below $10^{-4}$, and FP32 and FP64 evaluation accuracies differ by at most 0.14 percentage points. We observe no solver failures, NaNs, or infinite values. Thus, $\lambda>0$ guarantees positive definiteness, while the finite-precision behavior is supported empirically over the tested range rather than by a general guarantee of good conditioning.

\section{Limitations and Future Work}
\label{app:limitation}

CLP-DD is designed for the low-IPC frozen-backbone linear-probing regime, where the goal is to obtain an efficient synthetic set for a fixed pre-trained representation. Accordingly, its behavior is naturally influenced by the representation quality and feature geometry of the chosen backbone. This setting enables an exact closed-form inner solver and strong efficiency, but also makes the distilled images backbone-specific. The cross-backbone evaluation demonstrates limited transfer, but the substantial gap to matched-backbone performance confirms strong source-representation dependence; the primary setting remains adaptation to a given frozen encoder. The current evaluation transfers distilled images to a separately initialized, iteratively trained linear classifier, but does not establish robustness to substantially different probe losses, regularization choices, or nonlinear heads. In addition, while the sample-space solver is well-suited to low-IPC distillation, larger synthetic sets may benefit from primal, block-wise, or approximate solvers. CLP-DD is competitive with real-image selection under severe background shift, rather than consistently superior, and may inherit biases of the frozen representation. Future work will explore scalable solvers, multi-backbone objectives, broader downstream probes, and robustness-oriented outer objectives.

\section{Broader Impact}
\label{app:impact}

CLP-DD aims to reduce the computational and storage cost of adapting pre-trained vision models by replacing large training sets with compact synthetic sets. This can make frozen-backbone transfer learning more accessible in settings with limited computation, storage, or data-management resources. Since CLP-DD operates on fixed visual representations, the generated synthetic images reflect information encoded by the source data and the chosen backbone. As with prior dataset distillation methods, responsible use therefore requires appropriate validation for the target application, especially under distribution shift or bias-sensitive scenarios. CLP-DD does not introduce a generative model or a new data collection pipeline, and its primary impact is to improve the efficiency of existing transfer-learning workflows.

\section{Additional Distilled Images}
\label{app:additional_distilled_images}
This section provides additional qualitative visualizations of different datasets and backbones.

\begin{figure}[H]
    \centering
    \includegraphics[width=0.85\textwidth]{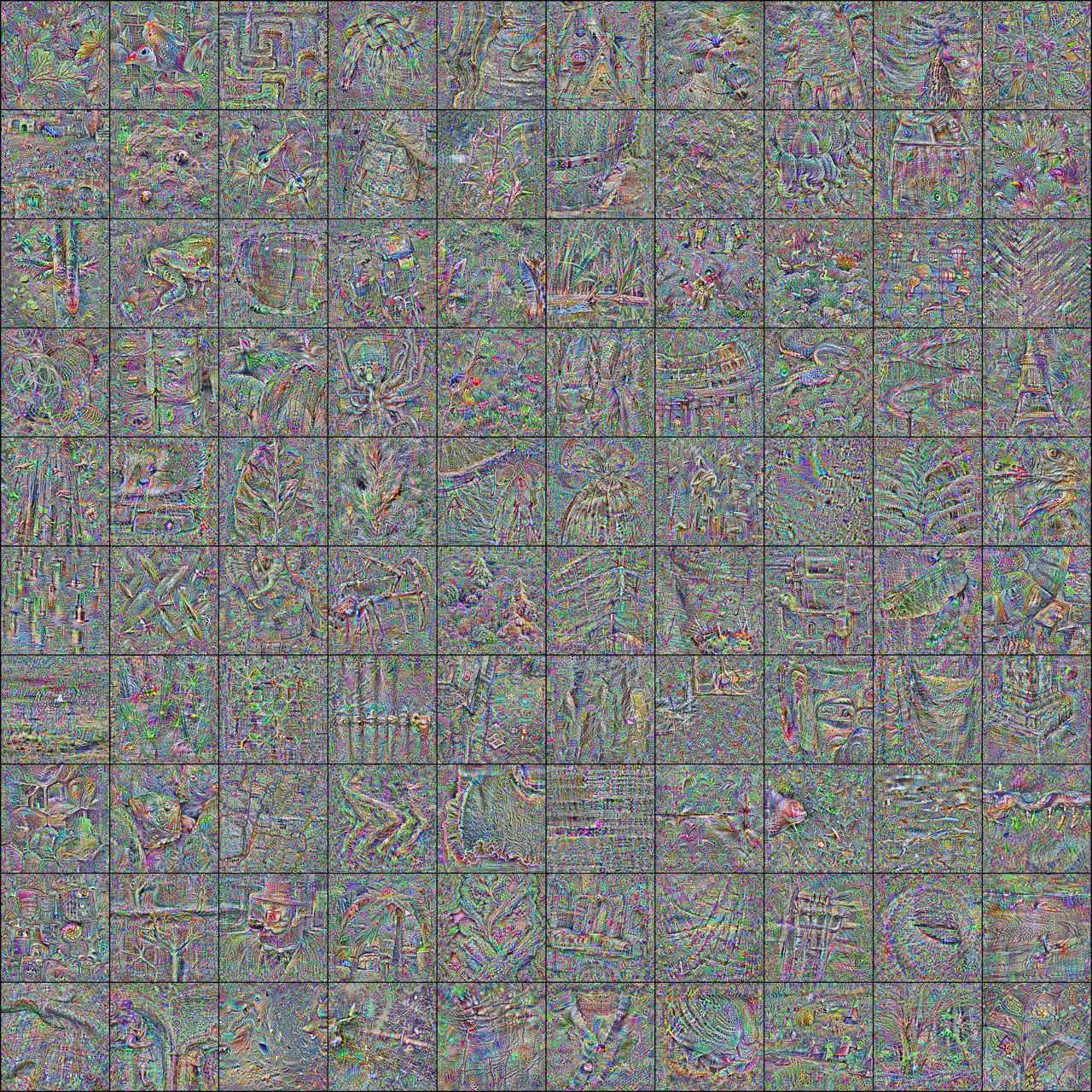}
    \caption{Additional examples of images distilled by CLP-DD on ImageNet-100 using the DINOv2 backbone.}
    \label{fig:app_CLP-DD_dino}
\end{figure}

\begin{figure}[H]
    \centering
    \includegraphics[width=0.85\textwidth]{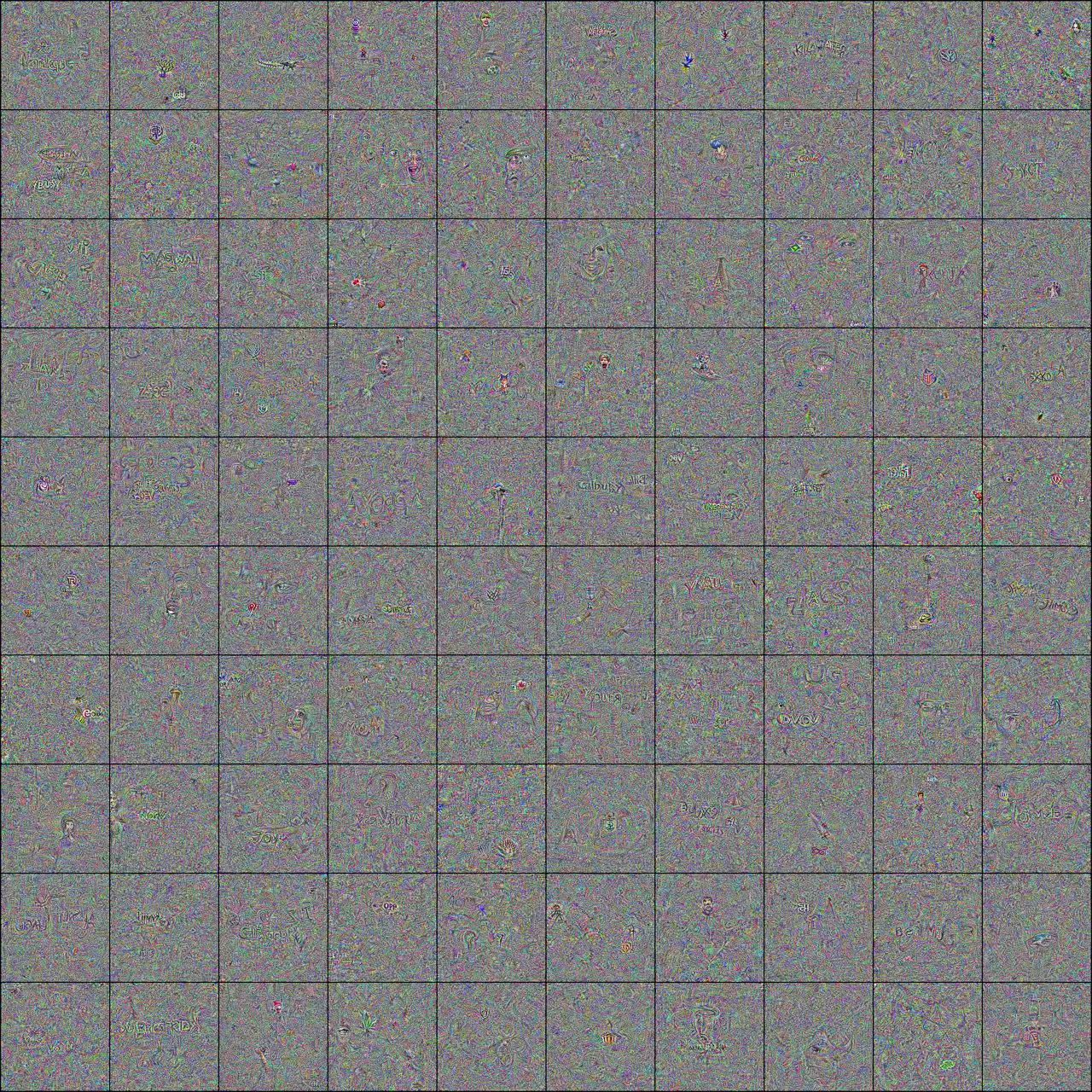}
    \caption{Additional examples of images distilled by CLP-DD on ImageNet-100 using the CLIP backbone.}
    \label{fig:app_CLP-DD_clip}
\end{figure}

\begin{figure}[H]
    \centering
    \includegraphics[width=0.85\textwidth]{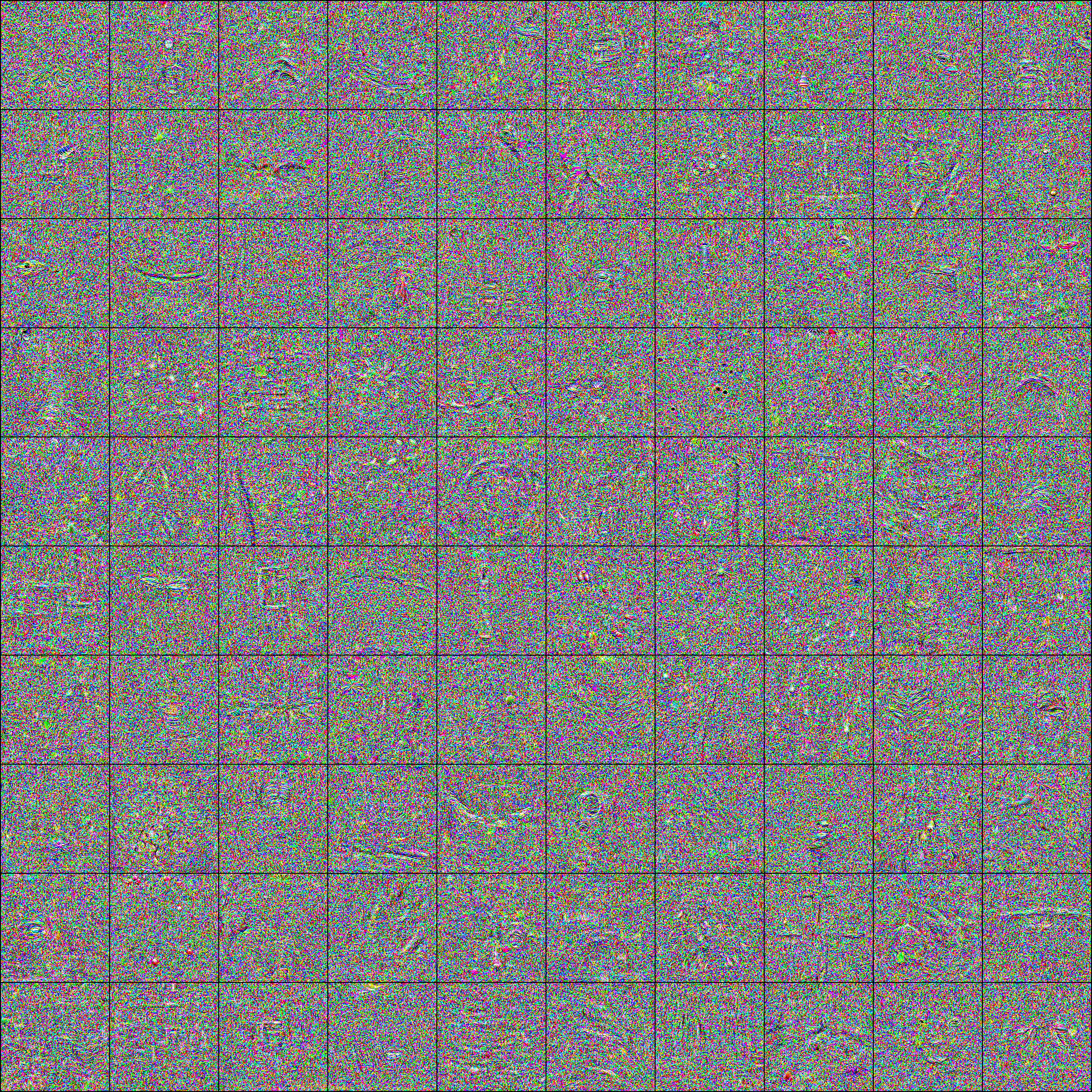}
    \caption{Additional examples of images distilled by CLP-DD on ImageNet-100 using the EVA-02 backbone.}
    \label{fig:app_CLP-DD_eva}
\end{figure}

\begin{figure}[H]
    \centering
    \includegraphics[width=0.85\textwidth]{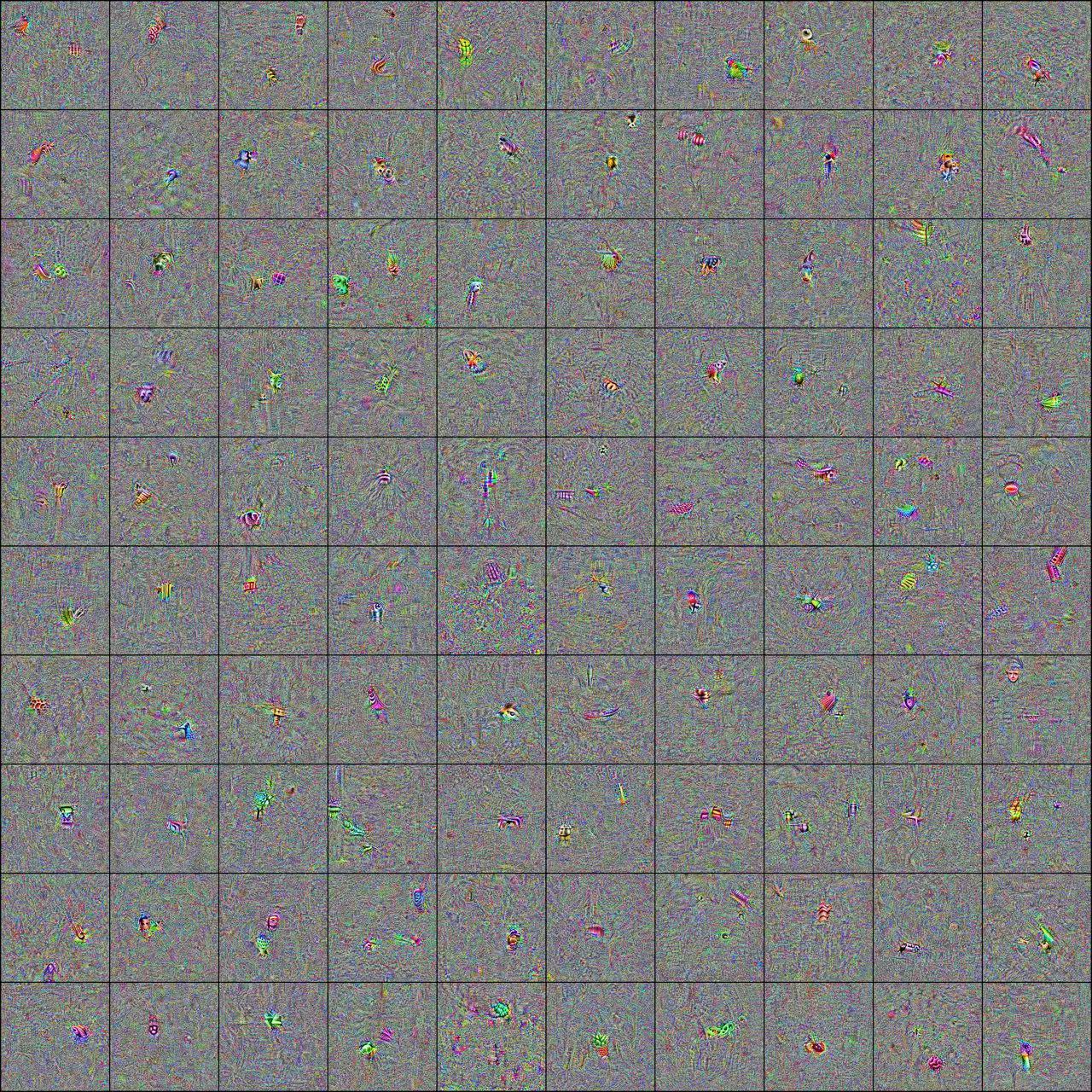}
    \caption{Additional examples of images distilled by CLP-DD on ImageNet-100 using the MoCo-v3 backbone.}
    \label{fig:app_CLP-DD_moco}
\end{figure}

\newpage
\section*{NeurIPS Paper Checklist}

\begin{enumerate}

\item {\bf Claims}
    \item[] Question: Do the main claims made in the abstract and introduction accurately reflect the paper's contributions and scope?
    \item[] Answer: \answerYes{} % Replace by \answerYes{}, \answerNo{}, or \answerNA{}.
    \item[] Justification: The abstract and introduction clearly state the main research claims. The theoretical basis is provided in Section~3 and Appendix~A, while the claimed experimental results in Section~4 support performance improvements.
    \item[] Guidelines:
    \begin{itemize}
        \item The answer \answerNA{} means that the abstract and introduction do not include the claims made in the paper.
        \item The abstract and/or introduction should clearly state the claims made, including the contributions made in the paper and important assumptions and limitations. A \answerNo{} or \answerNA{} answer to this question will not be perceived well by the reviewers. 
        \item The claims made should match theoretical and experimental results, and reflect how much the results can be expected to generalize to other settings. 
        \item It is fine to include aspirational goals as motivation as long as it is clear that these goals are not attained by the paper. 
    \end{itemize}

\item {\bf Limitations}
    \item[] Question: Does the paper discuss the limitations of the work performed by the authors?
    \item[] Answer: \answerYes{} % Replace by \answerYes{}, \answerNo{}, or \answerNA{}.
    \item[] Justification: We discuss the limitations of the proposed framework in the conclusion section and provide additional details in Appendix~\ref{app:limitation}.
    \item[] Guidelines:
    \begin{itemize}
        \item The answer \answerNA{} means that the paper has no limitation while the answer \answerNo{} means that the paper has limitations, but those are not discussed in the paper. 
        \item The authors are encouraged to create a separate ``Limitations'' section in their paper.
        \item The paper should point out any strong assumptions and how robust the results are to violations of these assumptions (e.g., independence assumptions, noiseless settings, model well-specification, asymptotic approximations only holding locally). The authors should reflect on how these assumptions might be violated in practice and what the implications would be.
        \item The authors should reflect on the scope of the claims made, e.g., if the approach was only tested on a few datasets or with a few runs. In general, empirical results often depend on implicit assumptions, which should be articulated.
        \item The authors should reflect on the factors that influence the performance of the approach. For example, a facial recognition algorithm may perform poorly when image resolution is low or images are taken in low lighting. Or a speech-to-text system might not be used reliably to provide closed captions for online lectures because it fails to handle technical jargon.
        \item The authors should discuss the computational efficiency of the proposed algorithms and how they scale with dataset size.
        \item If applicable, the authors should discuss possible limitations of their approach to address problems of privacy and fairness.
        \item While the authors might fear that complete honesty about limitations might be used by reviewers as grounds for rejection, a worse outcome might be that reviewers discover limitations that aren't acknowledged in the paper. The authors should use their best judgment and recognize that individual actions in favor of transparency play an important role in developing norms that preserve the integrity of the community. Reviewers will be specifically instructed to not penalize honesty concerning limitations.
    \end{itemize}

\item {\bf Theory assumptions and proofs}
    \item[] Question: For each theoretical result, does the paper provide the full set of assumptions and a complete (and correct) proof?
    \item[] Answer: \answerYes{} % Replace by \answerYes{}, \answerNo{}, or \answerNA{}.
    \item[] Justification: We provide complete explanations and proofs in Section~3 of the main text and Appendix~A.
    \item[] Guidelines:
    \begin{itemize}
        \item The answer \answerNA{} means that the paper does not include theoretical results. 
        \item All the theorems, formulas, and proofs in the paper should be numbered and cross-referenced.
        \item All assumptions should be clearly stated or referenced in the statement of any theorems.
        \item The proofs can either appear in the main paper or the supplemental material, but if they appear in the supplemental material, the authors are encouraged to provide a short proof sketch to provide intuition. 
        \item Inversely, any informal proof provided in the core of the paper should be complemented by formal proofs provided in appendix or supplemental material.
        \item Theorems and Lemmas that the proof relies upon should be properly referenced. 
    \end{itemize}

    \item {\bf Experimental result reproducibility}
    \item[] Question: Does the paper fully disclose all the information needed to reproduce the main experimental results of the paper to the extent that it affects the main claims and/or conclusions of the paper (regardless of whether the code and data are provided or not)?
    \item[] Answer: \answerYes{} % Replace by \answerYes{}, \answerNo{}, or \answerNA{}.
    \item[] Justification: The proposed framework is fully described in Section~3 of the manuscript, while Section~4 and Appendix~B present the detailed experimental configurations.
    \item[] Guidelines:
    \begin{itemize}
        \item The answer \answerNA{} means that the paper does not include experiments.
        \item If the paper includes experiments, a \answerNo{} answer to this question will not be perceived well by the reviewers: Making the paper reproducible is important, regardless of whether the code and data are provided or not.
        \item If the contribution is a dataset and\slash or model, the authors should describe the steps taken to make their results reproducible or verifiable. 
        \item Depending on the contribution, reproducibility can be accomplished in various ways. For example, if the contribution is a novel architecture, describing the architecture fully might suffice, or if the contribution is a specific model and empirical evaluation, it may be necessary to either make it possible for others to replicate the model with the same dataset, or provide access to the model. In general. releasing code and data is often one good way to accomplish this, but reproducibility can also be provided via detailed instructions for how to replicate the results, access to a hosted model (e.g., in the case of a large language model), releasing of a model checkpoint, or other means that are appropriate to the research performed.
        \item While NeurIPS does not require releasing code, the conference does require all submissions to provide some reasonable avenue for reproducibility, which may depend on the nature of the contribution. For example
        \begin{enumerate}
            \item If the contribution is primarily a new algorithm, the paper should make it clear how to reproduce that algorithm.
            \item If the contribution is primarily a new model architecture, the paper should describe the architecture clearly and fully.
            \item If the contribution is a new model (e.g., a large language model), then there should either be a way to access this model for reproducing the results or a way to reproduce the model (e.g., with an open-source dataset or instructions for how to construct the dataset).
            \item We recognize that reproducibility may be tricky in some cases, in which case authors are welcome to describe the particular way they provide for reproducibility. In the case of closed-source models, it may be that access to the model is limited in some way (e.g., to registered users), but it should be possible for other researchers to have some path to reproducing or verifying the results.
        \end{enumerate}
    \end{itemize}

\item {\bf Open access to data and code}
    \item[] Question: Does the paper provide open access to the data and code, with sufficient instructions to faithfully reproduce the main experimental results, as described in supplemental material?
    \item[] Answer: \answerNo{} % Replace by \answerYes{}, \answerNo{}, or \answerNA{}.
    \item[] Justification: We provide pseudocode and implementation details in the main paper and Appendix~\ref{app:implementation_details}. Complete reproduction scripts and configurations are not included in the paper package.
    \item[] Guidelines:
    \begin{itemize}
        \item The answer \answerNA{} means that paper does not include experiments requiring code.
        \item Please see the NeurIPS code and data submission guidelines (\url{https://neurips.cc/public/guides/CodeSubmissionPolicy}) for more details.
        \item While we encourage the release of code and data, we understand that this might not be possible, so \answerNo{} is an acceptable answer. Papers cannot be rejected simply for not including code, unless this is central to the contribution (e.g., for a new open-source benchmark).
        \item The instructions should contain the exact command and environment needed to run to reproduce the results. See the NeurIPS code and data submission guidelines (\url{https://neurips.cc/public/guides/CodeSubmissionPolicy}) for more details.
        \item The authors should provide instructions on data access and preparation, including how to access the raw data, preprocessed data, intermediate data, and generated data, etc.
        \item The authors should provide scripts to reproduce all experimental results for the new proposed method and baselines. If only a subset of experiments are reproducible, they should state which ones are omitted from the script and why.
        \item At submission time, to preserve anonymity, the authors should release anonymized versions (if applicable).
        \item Providing as much information as possible in supplemental material (appended to the paper) is recommended, but including URLs to data and code is permitted.
    \end{itemize}

\item {\bf Experimental setting/details}
    \item[] Question: Does the paper specify all the training and test details (e.g., data splits, hyperparameters, how they were chosen, type of optimizer) necessary to understand the results?
    \item[] Answer: \answerYes{} % Replace by \answerYes{}, \answerNo{}, or \answerNA{}.
    \item[] Justification: The detailed experimental setup, including hyperparameters, is fully described in Section~4 of the main text and Appendix~B.
    \item[] Guidelines:
    \begin{itemize}
        \item The answer \answerYes{} means that the paper does not include experiments.
        \item The experimental setting should be presented in the core of the paper to a level of detail that is necessary to appreciate the results and make sense of them.
        \item The full details can be provided either with the code, in appendix, or as supplemental material.
    \end{itemize}

\item {\bf Experiment statistical significance}
    \item[] Question: Does the paper report error bars suitably and correctly defined or other appropriate information about the statistical significance of the experiments?
    \item[] Answer: \answerYes{} % Replace by \answerYes{}, \answerNo{}, or \answerNA{}.
    \item[] Justification: Accuracy tables report mean and standard deviation over three repeated runs where available; profiling measurements and aggregate transfer averages are identified separately.
    \item[] Guidelines:
    \begin{itemize}
        \item The answer \answerNA{} means that the paper does not include experiments.
        \item The authors should answer \answerYes{} if the results are accompanied by error bars, confidence intervals, or statistical significance tests, at least for the experiments that support the main claims of the paper.
        \item The factors of variability that the error bars are capturing should be clearly stated (for example, train/test split, initialization, random drawing of some parameter, or overall run with given experimental conditions).
        \item The method for calculating the error bars should be explained (closed form formula, call to a library function, bootstrap, etc.)
        \item The assumptions made should be given (e.g., Normally distributed errors).
        \item It should be clear whether the error bar is the standard deviation or the standard error of the mean.
        \item It is OK to report 1-sigma error bars, but one should state it. The authors should preferably report a 2-sigma error bar than state that they have a 96\% CI, if the hypothesis of Normality of errors is not verified.
        \item For asymmetric distributions, the authors should be careful not to show in tables or figures symmetric error bars that would yield results that are out of range (e.g., negative error rates).
        \item If error bars are reported in tables or plots, the authors should explain in the text how they were calculated and reference the corresponding figures or tables in the text.
    \end{itemize}

\item {\bf Experiments compute resources}
    \item[] Question: For each experiment, does the paper provide sufficient information on the computer resources (type of compute workers, memory, time of execution) needed to reproduce the experiments?
    \item[] Answer: \answerYes{} % Replace by \answerYes{}, \answerNo{}, or \answerNA{}.
    \item[] Justification: In Section~4 and Appendix~B.3, we specify the computing resources used in our experiments.
    \item[] Guidelines:
    \begin{itemize}
        \item The answer \answerNA{} means that the paper does not include experiments.
        \item The paper should indicate the type of compute workers CPU or GPU, internal cluster, or cloud provider, including relevant memory and storage.
        \item The paper should provide the amount of compute required for each of the individual experimental runs as well as estimate the total compute. 
        \item The paper should disclose whether the full research project required more compute than the experiments reported in the paper (e.g., preliminary or failed experiments that didn't make it into the paper). 
    \end{itemize}
    
\item {\bf Code of ethics}
    \item[] Question: Does the research conducted in the paper conform, in every respect, with the NeurIPS Code of Ethics \url{https://neurips.cc/public/EthicsGuidelines}?
    \item[] Answer: \answerYes{} % Replace by \answerYes{}, \answerNo{}, or \answerNA{}.
    \item[] Justification: The authors have carefully reviewed the NeurIPS Code of Ethics and conducted the research in accordance with it.
    \item[] Guidelines:
    \begin{itemize}
        \item The answer \answerNA{} means that the authors have not reviewed the NeurIPS Code of Ethics.
        \item If the authors answer \answerNo, they should explain the special circumstances that require a deviation from the Code of Ethics.
        \item The authors should make sure to preserve anonymity (e.g., if there is a special consideration due to laws or regulations in their jurisdiction).
    \end{itemize}

\item {\bf Broader impacts}
    \item[] Question: Does the paper discuss both potential positive societal impacts and negative societal impacts of the work performed?
    \item[] Answer: \answerYes{} % Replace by \answerYes{}, \answerNo{}, or \answerNA{}.
    \item[] Justification: In this study, we propose a novel perspective on dataset distillation and discuss its broader impact in Appendix~\ref{app:impact}.
\begin{itemize}
    \item reducing computational resource requirements for deep learning.

    \item decreasing energy consumption associated with large-scale training.
\end{itemize}
 
    \item[] Guidelines:
    \begin{itemize}
        \item The answer \answerNA{} means that there is no societal impact of the work performed.
        \item If the authors answer \answerNA{} or \answerNo, they should explain why their work has no societal impact or why the paper does not address societal impact.
        \item Examples of negative societal impacts include potential malicious or unintended uses (e.g., disinformation, generating fake profiles, surveillance), fairness considerations (e.g., deployment of technologies that could make decisions that unfairly impact specific groups), privacy considerations, and security considerations.
        \item The conference expects that many papers will be foundational research and not tied to particular applications, let alone deployments. However, if there is a direct path to any negative applications, the authors should point it out. For example, it is legitimate to point out that an improvement in the quality of generative models could be used to generate Deepfakes for disinformation. On the other hand, it is not needed to point out that a generic algorithm for optimizing neural networks could enable people to train models that generate Deepfakes faster.
        \item The authors should consider possible harms that could arise when the technology is being used as intended and functioning correctly, harms that could arise when the technology is being used as intended but gives incorrect results, and harms following from (intentional or unintentional) misuse of the technology.
        \item If there are negative societal impacts, the authors could also discuss possible mitigation strategies (e.g., gated release of models, providing defenses in addition to attacks, mechanisms for monitoring misuse, mechanisms to monitor how a system learns from feedback over time, improving the efficiency and accessibility of ML).
    \end{itemize}
    
\item {\bf Safeguards}
    \item[] Question: Does the paper describe safeguards that have been put in place for responsible release of data or models that have a high risk for misuse (e.g., pre-trained language models, image generators, or scraped datasets)?
    \item[] Answer: \answerNA{} % Replace by \answerYes{}, \answerNo{}, or \answerNA{}.
    \item[] Justification: This study proposes a dataset distillation framework whose risks are not substantially higher than those of prior dataset distillation methods. The framework does not include a generative component, and the potential for misuse of the research results is limited. Therefore, we consider this item not applicable to our study.
    \item[] Guidelines:
    \begin{itemize}
        \item The answer \answerNA{} means that the paper poses no such risks.
        \item Released models that have a high risk for misuse or dual-use should be released with necessary safeguards to allow for controlled use of the model, for example by requiring that users adhere to usage guidelines or restrictions to access the model or implementing safety filters. 
        \item Datasets that have been scraped from the Internet could pose safety risks. The authors should describe how they avoided releasing unsafe images.
        \item We recognize that providing effective safeguards is challenging, and many papers do not require this, but we encourage authors to take this into account and make a best faith effort.
    \end{itemize}

\item {\bf Licenses for existing assets}
    \item[] Question: Are the creators or original owners of assets (e.g., code, data, models), used in the paper, properly credited and are the license and terms of use explicitly mentioned and properly respected?
    \item[] Answer: \answerYes{} % Replace by \answerYes{}, \answerNo{}, or \answerNA{}.
    \item[] Justification: The datasets, models, and other resources used in this paper have been properly cited.
    \item[] Guidelines:
    \begin{itemize}
        \item The answer \answerNA{} means that the paper does not use existing assets.
        \item The authors should cite the original paper that produced the code package or dataset.
        \item The authors should state which version of the asset is used and, if possible, include a URL.
        \item The name of the license (e.g., CC-BY 4.0) should be included for each asset.
        \item For scraped data from a particular source (e.g., website), the copyright and terms of service of that source should be provided.
        \item If assets are released, the license, copyright information, and terms of use in the package should be provided. For popular datasets, \url{paperswithcode.com/datasets} has curated licenses for some datasets. Their licensing guide can help determine the license of a dataset.
        \item For existing datasets that are re-packaged, both the original license and the license of the derived asset (if it has changed) should be provided.
        \item If this information is not available online, the authors are encouraged to reach out to the asset's creators.
    \end{itemize}

\item {\bf New assets}
    \item[] Question: Are new assets introduced in the paper well documented and is the documentation provided alongside the assets?
    \item[] Answer: \answerNA{} % Replace by \answerYes{}, \answerNo{}, or \answerNA{}.
    \item[] Justification: This paper does not introduce a new benchmark dataset. Implementation details are provided in Appendix~\ref{app:implementation_details}.
    \item[] Guidelines:
    \begin{itemize}
        \item The answer \answerNA{} means that the paper does not release new assets.
        \item Researchers should communicate the details of the dataset\slash code\slash model as part of their submissions via structured templates. This includes details about training, license, limitations, etc. 
        \item The paper should discuss whether and how consent was obtained from people whose asset is used.
        \item At submission time, remember to anonymize your assets (if applicable). You can either create an anonymized URL or include an anonymized zip file.
    \end{itemize}

\item {\bf Crowdsourcing and research with human subjects}
    \item[] Question: For crowdsourcing experiments and research with human subjects, does the paper include the full text of instructions given to participants and screenshots, if applicable, as well as details about compensation (if any)? 
    \item[] Answer: \answerNA{} % Replace by \answerYes{}, \answerNo{}, or \answerNA{}.
    \item[] Justification: This study does not involve crowdsourcing experiments or research with human subjects.
    \item[] Guidelines:
    \begin{itemize}
        \item The answer \answerNA{} means that the paper does not involve crowdsourcing nor research with human subjects.
        \item Including this information in the supplemental material is fine, but if the main contribution of the paper involves human subjects, then as much detail as possible should be included in the main paper. 
        \item According to the NeurIPS Code of Ethics, workers involved in data collection, curation, or other labor should be paid at least the minimum wage in the country of the data collector. 
    \end{itemize}

\item {\bf Institutional review board (IRB) approvals or equivalent for research with human subjects}
    \item[] Question: Does the paper describe potential risks incurred by study participants, whether such risks were disclosed to the subjects, and whether Institutional Review Board (IRB) approvals (or an equivalent approval/review based on the requirements of your country or institution) were obtained?
    \item[] Answer: \answerNA{} % Replace by \answerYes{}, \answerNo{}, or \answerNA{}.
    \item[] Justification: This study does not involve human-subject research or crowdsourcing and therefore does not require IRB approval.
    \item[] Guidelines:
    \begin{itemize}
        \item The answer \answerNA{} means that the paper does not involve crowdsourcing nor research with human subjects.
        \item Depending on the country in which research is conducted, IRB approval (or equivalent) may be required for any human subjects research. If you obtained IRB approval, you should clearly state this in the paper. 
        \item We recognize that the procedures for this may vary significantly between institutions and locations, and we expect authors to adhere to the NeurIPS Code of Ethics and the guidelines for their institution. 
        \item For initial submissions, do not include any information that would break anonymity (if applicable), such as the institution conducting the review.
    \end{itemize}

\item {\bf Declaration of LLM usage}
    \item[] Question: Does the paper describe the usage of LLMs if it is an important, original, or non-standard component of the core methods in this research? Note that if the LLM is used only for writing, editing, or formatting purposes and does \emph{not} impact the core methodology, scientific rigor, or originality of the research, declaration is not required.
    %this research? 
    \item[] Answer: \answerNA{} % Replace by \answerYes{}, \answerNo{}, or \answerNA{}.
    \item[] Justification: This study does not involve large language models (LLMs) in its core methodology, data processing, or experimental design. The research is based on dataset distillation techniques without any LLM-related components. Therefore, no declaration of LLM usage is required for this submission.
    \item[] Guidelines:
    \begin{itemize}
        \item The answer \answerNA{} means that the core method development in this research does not involve LLMs as any important, original, or non-standard components.
        \item Please refer to our LLM policy in the NeurIPS handbook for what should or should not be described.
    \end{itemize}

\end{enumerate}

\end{document}